\documentclass{article}

\usepackage[numbers]{natbib}

    \usepackage[preprint]{neurips_2020}

\usepackage[utf8]{inputenc} % allow utf-8 input
\usepackage[T1]{fontenc}    % use 8-bit T1 fonts
\usepackage{hyperref}       % hyperlinks
\usepackage{url}            % simple URL typesetting
\usepackage{booktabs}       % professional-quality tables
\usepackage{amsfonts}       % blackboard math symbols
\usepackage{nicefrac}       % compact symbols for 1/2, etc.
\usepackage{microtype}      % microtypography

\usepackage{graphicx}
\usepackage{dcolumn}
\usepackage{subcaption}
\usepackage{amsmath}
\usepackage{amssymb}
\usepackage{multirow}
\usepackage{multicol}
\usepackage{enumitem}
\usepackage{bm}
\usepackage{algorithm}
\usepackage{algpseudocode}
\usepackage{wrapfig,lipsum,booktabs}
\usepackage{adjustbox}

\def \zz {\mathbf{z}}

\def \zz {\mathbf{z}}

\DeclareMathOperator{\sign}{sign}
\newcommand{\norm}[1]{\left\lVert#1\right\rVert}
\usepackage{xspace}
\title{Imbalanced Gradients: A Subtle Cause of Overestimated Adversarial Robustness}

\author{%
Xingjun Ma\footnotemark[1]\\
   School of Computer Science \\
   Fudan University, China\\
   \texttt{xingjunma@fudan.edu.cn} \\
   \And
  Linxi Jiang\footnotemark[1] \\
  School of Computer Science \\
  Fudan University, China \\
  \texttt{lxjiang18@fudan.edu.cn} \\
   \And
   Hanxun Huang \footnotemark[2]\\
   School of Computing and Information Systems \\
   The University of Melbourne, Australia \\
   \texttt{hanxunh@student.unimelb.edu.au} \\
   \And
   Zejia Weng \\
   School of Computer Science \\
  Fudan University, China \\
   \texttt{zjweng16@fudan.edu.cn} \\
   \And
   James Bailey \\
   School of Computing and Information Systems \\
   The University of Melbourne, Australia \\
   \texttt{baileyj@unimelb.edu.au} \\
   \And
   Yu-Gang Jiang\footnotemark[2] \\
   School of Computer Science \\
  Fudan University, China \\
   \texttt{ygj@fudan.edu.cn} \\
}

\begin{document}

\maketitle

\renewcommand{\thefootnote}{\fnsymbol{footnote}} %将脚注符号设置为fnsymbol类型，即特殊符号表示
\footnotetext[1]{Equal contribution.
} %对应脚注[1]
\footnotetext[2]{Corresponding authors. } %对应脚注[2]

\begin{abstract}
Evaluating the robustness of a defense model is a challenging task in adversarial robustness research.
Obfuscated gradients have previously been found to exist in many defense methods and cause a false signal of robustness.
In this paper, we identify a more subtle situation called \emph{Imbalanced Gradients} that can also cause overestimated adversarial robustness. 
The phenomenon of imbalanced gradients occurs when the gradient of one term of the margin loss dominates and pushes the attack towards to a suboptimal direction.
To exploit imbalanced gradients, we formulate a \emph{Margin Decomposition (MD)} attack that decomposes a margin loss into individual terms and then explores the attackability of these terms separately via a two-stage process. We also propose a multi-targeted and ensemble version of our MD attack.
By investigating 24 defense models proposed since 2018, we find that 11 models are susceptible to a certain degree of imbalanced gradients and our MD attack can decrease their robustness evaluated by the best standalone baseline attack by more than 1\%.
We also provide an in-depth investigation on the likely causes of imbalanced gradients and effective countermeasures. Our code is available at \url{https://github.com/HanxunH/MDAttack}.
\end{abstract}

\section{Introduction}\label{sec:introduction}
Deep neural networks (DNNs) are vulnerable to adversarial examples, which are input instances crafted by adding small adversarial perturbations to natural examples. Adversarial examples can fool DNNs into making false predictions with high confidence, and transfer across different models \citep{szegedy2013intriguing,goodfellow2014fgsm}. A number of defenses have been proposed to overcome this vulnerability. However, a concerning fact is that many defenses have been quickly shown to have undergone incorrect or incomplete evaluation \citep{athalye2018obfuscated,logan2018evalalp,carlini2019evaluating,tramer2020adaptive,croce2020reliable}. One common pitfall in adversarial robustness evaluation is the phenomenon of gradient masking \citep{Papernot2017practical,Tramer2018ensembleadvtrain} or obfuscated gradients \citep{athalye2018obfuscated},  leading to weak or unsuccessful attacks and false signals of robustness.
To demonstrate ``real" robustness, newly proposed defenses claim robustness based on results of white-box attacks such as Projected Gradient Decent (PGD) attack \citep{madry2017towards} and AutoAttack \citep{croce2020reliable,croce2020robustbench}, and demonstrate that they are not a result of obfuscated gradients.
% One natural question to ask here is that ``Can these attacks produce reliable evaluation when there are no obfuscated gradients"? 
In this work, we show that the robustness may still be overestimated even when there are no obfuscated gradients. Specifically, we identify a subtle situation called \emph{Imbalanced Gradients} that exists in several recent defense models and can cause highly overestimated robustness.

Imbalanced gradients is a new type of gradient masking effect where the gradient of one loss term dominates that of other terms. This causes the attack to move toward a suboptimal direction. Different from obfuscated gradients, imbalanced gradients are more subtle and are not detectable by the detection methods used for obfuscated gradients.
To exploit imbalanced gradients, we propose a novel attack named \emph{Margin Decomposition (MD)} attack that decomposes the margin loss into two separate terms and then exploits the attackability of these terms via a two-stage attacking process. We also derive the MultiTargeted \citep{gowal2019multitargeted} and ensemble variants of MD attack, following AutoAttack. By examining the robustness of 24 adversarial training based defense models proposed since 2018. We find that 11 of them are susceptible to imbalanced gradients to a certain extent, and their robustness evaluated by the best baseline standalone attack drops by more than 1\% against our MD attack.
Our key contributions are:
\begin{itemize}
    \item We identify a new type of effect called {\em imbalanced gradients}, which can cause overestimated adversarial robustness and cannot be detected by detection methods for obfuscated gradients. Especially, we highlight that label smoothing is one of the major causes of imbalanced gradients.
    
    \item We propose \emph{Margin Decomposition (MD)} attacks to exploit imbalanced gradients. MD leverages the attackability of the individual terms in the margin loss in a two-stage attacking process. We also introduce the MultiTargeted and ensemble variants of MD.
    
    \item We conduct extensive evaluations on 24 state-of-the-art defense models and find that 11 of them suffer from imbalanced gradients to some extent and their robustness evaluated by the best standalone attack drops by more than 1\% against our MD attack. Our MD Ensemble (MDE) attack exceeds state-of-the-art attack AutoAttack on 16/20 defense models on CIFAR-10.  Our MD attack alone can outperform AutoAttack in evaluating the adversarial robustness of vision transformer and ResNet-50 models trained on ImageNet.
\end{itemize}

\section{Related Work}
We denote a clean sample by $\bm{x}$, its class by $y \in \{1, \cdots, C\}$ with $C$ the number of classes, and a DNN classifier by $f$. The probability of $\bm{x}$ being in the $i$-th class is computed as
$\bm{p}_i(\bm{x})=e^{\bm{z}_i}/\sum_{j=1}^C e^{\bm{z}_{j}}$, where $\bm{z}_i$ is the logits for the $i$-th class. The goal of an adversarial attack is to find an adversarial example $\bm{x}_{adv}$ that can fool the model into making a false prediction (e.g., $f(\bm{x}_{adv}) \neq y$), and is typically restricted to be within a small $\epsilon$-ball around the original sample $\bm{x}$ (e.g., $\norm{\bm{x}_{adv} - \bm{x}}_\infty \leq \epsilon$).

\noindent\textbf{Adversarial Attack.}
Adversarial examples can be crafted by maximizing a classification loss $\ell$ by one or multiple steps of adversarial perturbations.
For example, the one-step Fast Gradient Sign Method (FGSM) \citep{goodfellow2014fgsm} and the iterative FGSM (I-FGSM) attack \citep{kurakin2016adversarial}. Projected Gradient Descent (PGD) \citep{madry2017towards} attack is another iterative method that projects the perturbation back onto the $\epsilon$-ball centered at $\bm{x}$ when it goes beyond.
Carlini and Wagner (CW) \citep{carlini2017towards} attack generates adversarial examples via an optimization framework. There also exist other attacks such as Frank-Wolfe attack \citep{chen2018frank}, distributionally adversarial attack \citep{zheng2019distributionally} and elastic-net attacks \citep{chen2018ead}. In earlier literature, the most commonly used attacks for robustness evaluations are FGSM, PGD, and CW attacks.

Several recent attacks have been proposed to produce more accurate robustness evaluations than PGD.
This includes Fast Adaptive Boundary Attack (FAB) \citep{croce2019minimally}, MultiTargeted (MT) attack \citep{gowal2019multitargeted}, Output Diversified Initialization (ODI) \citep{tashiro2020ods}, and AutoAttack (AA) \citep{croce2020reliable}.
FAB finds the minimal perturbation necessary to change the class of a given input.
MT \citep{gowal2019multitargeted} is a PGD-based attack with multiple restarts and picks a new target class at each restart. ODI provides a more effective initialization strategy with diversified logits. AA attack is a parameter-free ensemble of four attacks: FAB, two Auto-PGD attacks, and the black-box Square Attack \citep{andriushchenko2019square}. AA has demonstrated to be one of the state-of-the-art attacks to date, according to the RobustBench \citep{croce2020robustbench}.

\noindent\textbf{Adversarial Loss.}
Many attacks use Cross Entropy (CE) as the adversarial loss: $\ell_{ce}(\bm{x}, y) = -\log \bm{p}_y$.
The other commonly used adversarial loss is the margin loss \citep{carlini2017towards}: $\ell_{margin}(\bm{x}, y) = \bm{z}_{max} - \bm{z}_{y}$,
with $\bm{z}_{max} = \max_{i \neq y}\bm{z}_i$. Shown in  \citep{gowal2019multitargeted}, CE can be written in a margin form (e.g., $\ell_{ce}(\bm{x}, y)=\log(\sum^{C}_{i=1} e^{\bm{z}_i}) - \bm{z}_{y}$), and in most cases, they are both effective. While FGSM and PGD attacks use the CE loss, CW and several recent attacks such as MT and ODI adopt the margin loss.
AA has one PGD variant using the CE loss and the other PGD variant using the Difference of Logits Ratio (DLR) loss. DLR can be regarded as a ``relative margin'' loss.
In this paper, we identify a new effect that causes overestimated adversarial robustness from the margin loss perspective and propose new attacks by decomposing the margin loss.

\noindent\textbf{Adversarial Defense.}
In response to the threat of adversarial attacks, many defenses have been proposed such as defensive distillation \citep{papernot2016distillation}, feature/subspace analysis \citep{xu2017feature, ma2018characterizing}, denoising techniques \citep{guo2018countering,liao2018defense,samangouei2018defensegan}, robust regularization \citep{gu2014towards, tramer2017ensemble, ross2018improving}, model compression \citep{liu2018security,dascompression,rakin2018defend} and adversarial training \citep{goodfellow2014fgsm,madry2017towards}. Among them, adversarial training via robust min-max optimization has been found to be the most effective approach \citep{athalye2018obfuscated}.
The standard adversarial training (SAT) \citep{madry2017towards} trains models on adversarial examples generated via the PGD attack.
Dynamic adversarial training (Dynamic) \citep{wang2019convergence} trains on adversarial examples with gradually increased convergence quality.
Max-Margin Adversarial training (MMA) \citep{ding2018mma} trains on adversarial examples with gradually increased margin (e.g., the perturbation bound $\epsilon$).
Jacobian Adversarially Regularized Networks (JARN) adversarially regularize the Jacobian matrices, and can be combined with 1-step adversarial training (JARN-AT1) to gain additional robustness \citep{chan2020jacobian}.
Sensible adversarial training (Sense) \citep{kim2020sensible} trains on loss-sensible adversarial examples (perturbation stops when loss exceeds certain threshold). 
Adversarial Training with Pre-Training (AT-PT) \citep{hendrycks2019using} uses pre-training to improve model robustness. Adversarial Training with Early Stopping (AT-ES) \citep{rice2020overfitting} suggests the use of early stopping to avoid the robust overfitting of adversarial training.
Bilateral Adversarial Training (Bilateral) \citep{wang2019bilateral} trains on PGD adversarial examples with adversarially perturbed labels. Adversarial Interpolation (Adv-Interp) training \citep{zhang2020adversarial} trains on adversarial examples generated under an adversarial interpolation scheme with adversarial labels.
Feature Scattering-based (FeaScatter) adversarial training \citep{zhang2019featurescatter} crafts adversarial examples using latent space feature scattering, then trains on these examples with label smoothing. 
TRADES \citep{zhang2019trades} replaces the CE loss of SAT by the KL divergence for a better trade-off between robustness and natural accuracy. Based on TRADES, RST \citep{carmon2019unlabel} and UAT \citep{stanforth2019labels} improve robustness by training with $10\times$ more unlabeled data. Misclassification Aware adveRsarial Training (MART) \citep{wang2020improving} further improves the above three methods with a misclassification aware loss function.
Adversarial Weight Perturbation (AWP) \citep{wu2020adversarial} perturbs inputs and model weights alternatively during adversarial training to improve robust generalization.
Channel-wise Activation Suppressing (CAS) robustifies the intermediate layers of DNNs via an auxiliary channel suppressing module \citep{bai2020improving}.
There are also recent works on robust neural architectures \citep{shao2021adversarial,du2021learning,tang2021robustart,huang2021exploring} and adversarial robustness distillation \citep{goldblum2020adversarially,zhu2021reliable,zi2021revisiting}.
We will discuss and evaluate a set of the above adversarial training-based defenses in Section \ref{sec:experiments}.

\noindent\textbf{Evaluation of Adversarial Robustness.}
Adversarial robustness requires careful and rigorous evaluation.
Many defenses that perform incomplete evaluation are quickly broken by new attacks. Several evaluation pitfalls have been identified as needing to be avoided for reliable robustness evaluation \citep{carlini2019evaluating}.
Although several general principles have been suggested around the regular attacks such as PGD \citep{carlini2019evaluating}, there are scenarios where these attacks may give unreliable robustness evaluation.
Gradient masking \citep{Tramer2018ensembleadvtrain,Papernot2017practical} is a common effect that blocks the attack by hiding useful gradient information. Obfuscated gradients \citep{athalye2018obfuscated}, a type of gradient masking, have been exploited (unintentionally) by many defense methods to cause an overly optimistic evaluation of robustness. Obfuscated gradients exist in different forms such as non-differentiable gradients, stochastic gradients, or vanishing/exploding gradients. Such defenses have all been successfully circumvented by adaptive attacks in \citep{athalye2018obfuscated,carlini2019evaluating,tramer2020adaptive}. Gradient-free attacks such as SPSA \citep{spall1992multivariate} have also been used to identify obscured models \citep{Uesato2018spsa}. In this paper, we identify a more subtle situation called \emph{Imbalanced Gradients}, which also causes overestimated robustness, but is different from obfuscated gradients.

\section{Imbalanced Gradients and Robustness Evaluation} \label{sec:toy}
We first give a toy example of imbalanced gradients and show how regular attacks can fail in such a situation. We then empirically verify their existence in deep neural networks, particularly for some adversarially-trained models. Finally, we propose the Margin Decomposition attacks to exploit the imbalanced gradients. Since CE and margin loss are the two commonly used loss functions for adversarial attacks and CE can be written in a margin form \citep{gowal2019multitargeted}, here we focus on the margin loss to present the phenomenon of imbalanced gradients.

\noindent\textbf{Imbalanced Gradients.} The gradient of the margin loss (e.g., $\ell_{margin}(\bm{x}, y) = \bm{z}_{max} - \bm{z}_{y}$) is the combination of the gradients of its two individual terms (e.g., $\nabla_{\bm{x}} (\bm{z}_{max} - \bm{z}_y) = \nabla_{\bm{x}} \bm{z}_{max} + \nabla_{\bm{x}} (-\bm{z}_y)$). 
\emph{Imbalanced Gradients} is the situation where the gradient of one loss term dominates that of other term(s), pushing the attack towards a suboptimal direction. 

\begin{wrapfigure}{R}{6cm}
\vskip -0.5in
\begin{center}
        \includegraphics[width=1.0\linewidth]{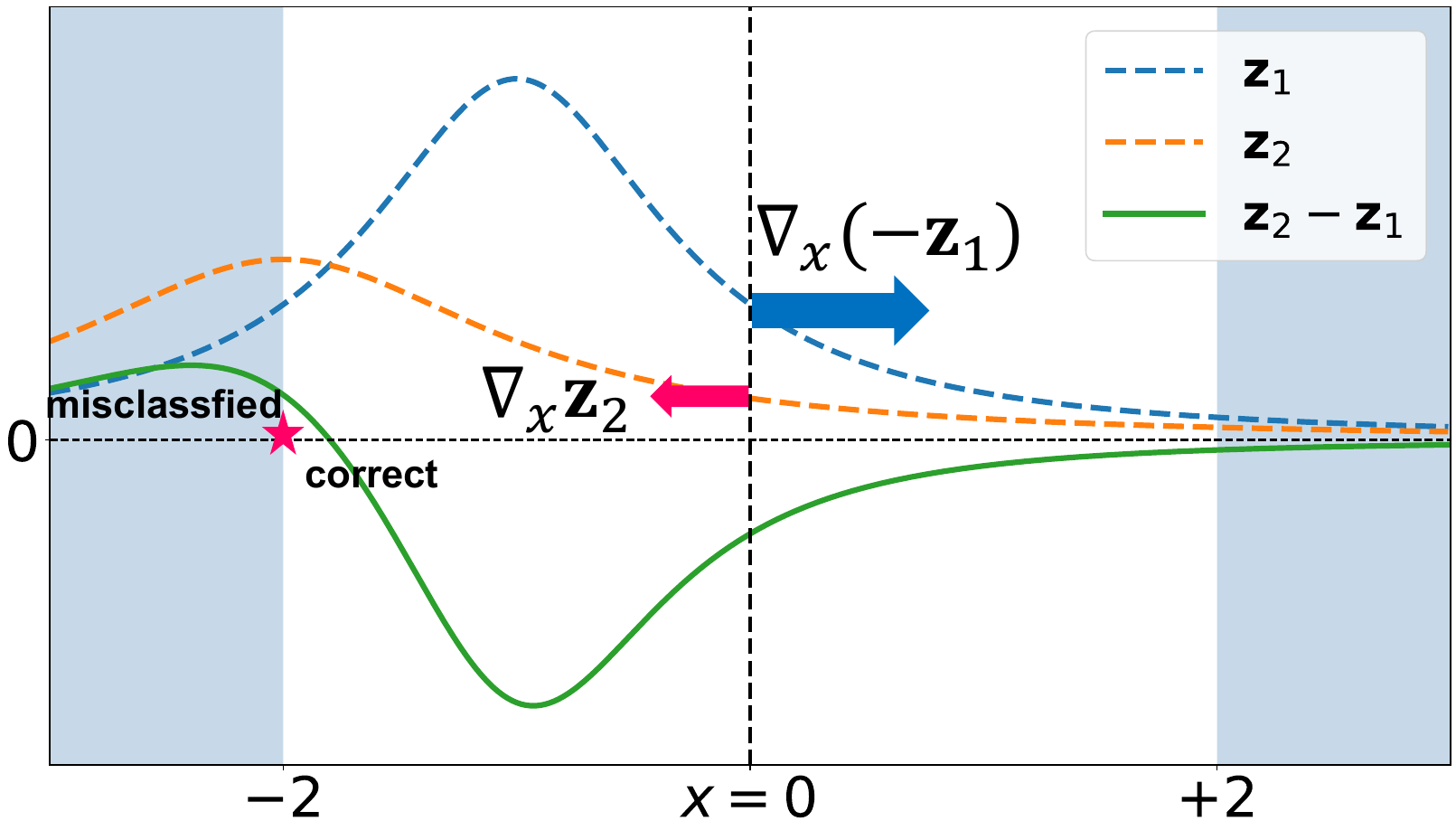}
    \end{center}
    \caption{A toy illustration of \emph{imbalanced gradients} at $x=0$: the gradient of margin loss ($\zz_2 - \zz_1$) is dominated by its $- \zz_1$ term, pointing to a suboptimal attack direction towards +2, where $x$ is still correctly classified.}
    \label{fig:toy-a}
\vskip -0.1in
\end{wrapfigure}

\noindent\textbf{Toy Example.}
Consider a one-dimensional classification task and a binary classifier with two outputs $\bm{z}_1$ and $\bm{z}_2$ (like logits of a DNN),  Fig. \ref{fig:toy-a} illustrates the distributions of $\bm{z}_1$, $\bm{z}_2$ and $\bm{z}_2 - \bm{z}_1$ around $x=0$. The classifier predicts class 1 when $\bm{z}_1 \geq \bm{z}_2$, otherwise class 2. We consider an input at $x = 0$ with correct prediction $y=1$, and a maximum perturbation constraint $\epsilon = 2$ (e.g., perturbation $\delta \in [-2, +2]$).
The attack is successful if and only if $\bm{z}_2 > \bm{z}_1$. 
In this example, imbalanced gradients occur at $x=0$, where the gradients of the two terms $\nabla_{x} \bm{z}_2$ and $\nabla_{x} (- \bm{z}_1)$ have opposite directions, and the attack is dominated by the $\bm{z}_1$ term as $\nabla_{x} (- \bm{z}_1)$ is significantly larger than $\nabla_{x} \bm{z}_2$. Thus, attacking $x$ with the margin loss will converge to +2, where the sample is still correctly classified. However, for a successful attack, $x$ should be perturbed towards -2. In this particular scenario, the gradient $\nabla_{x} \bm{z}_2 < 0$ alone can provide the most effective attack direction. Note that this toy example was motivated by the loss landscape of DNNs when imbalanced gradients occur.

\subsection{Imbalanced Gradients in DNNs}
The situation can be extremely complex for DNNs with high-dimensional inputs, as imbalanced gradients can occur at each input dimension. 
It thus requires a metric to quantitatively measure the degree of gradient imbalance. Here, we propose such a metric named \emph{Gradient Imbalance Ratio} (GIR) to measure the imbalance ratio for a single input $\bm{x}$, which can then be averaged over multiple inputs to produce the imbalance ratio for the entire model.

% \vspace{0.05in}
\noindent\textbf{Definition of GIR.} To measure the imbalance ratio, we focus on the input dimensions that are dominated by one loss term. An input dimension $x_i$ is dominated by a loss term (e.g., $\bm{z}_{max}$) means that 1) the gradients of loss terms at $x_i$ have different directions ($\nabla_{x_{i}} \bm{z}_{max} \cdot \nabla_{x_{i}} (-\bm{z}_{y}) < 0$), and 2) the gradient of the dominant term is larger (e.g., $\lvert\nabla_{x_{i}}\bm{z}_{max}\rvert > \lvert\nabla_{x_{i}} (-\bm{z}_y)\rvert$). 
According to the dominant term, we can split these dimensions into two subsets $\bm{x}_{s_1}$ and $\bm{x}_{s_2}$ where $\bm{x}_{s_1}$ are dominated by the $\bm{z}_{max}$ term, while $\bm{x}_{s_2}$ are dominated by the $-\bm{z}_{y}$ term. The overall dominance effect of each loss term can be formulated as $r_1=\norm{\nabla_{\bm{x}_{s_1}}(\bm{z}_{max} - \bm{z}_y)}_1$ and $r_2=\norm{\nabla_{\bm{x}_{s_2}}(\bm{z}_{max} - \bm{z}_y)}_1$. Here, we use the  $L_1$-norms instead of $L_0$-norms (i.e., the number of dominated dimensions) to also take into consideration the gradient magnitude. To keep the ratio larger than 1, GIR is computed as: 

\begin{equation}
    GIR = \max\{\frac{r_1}{r_2}, \frac{r_2}{r_1}\}.
\end{equation}

\begin{figure*}[!t]
    \vskip -0.1in
    \centering
    \subcaptionbox{ \label{fig:mir}}{\includegraphics[width=0.33\textwidth]{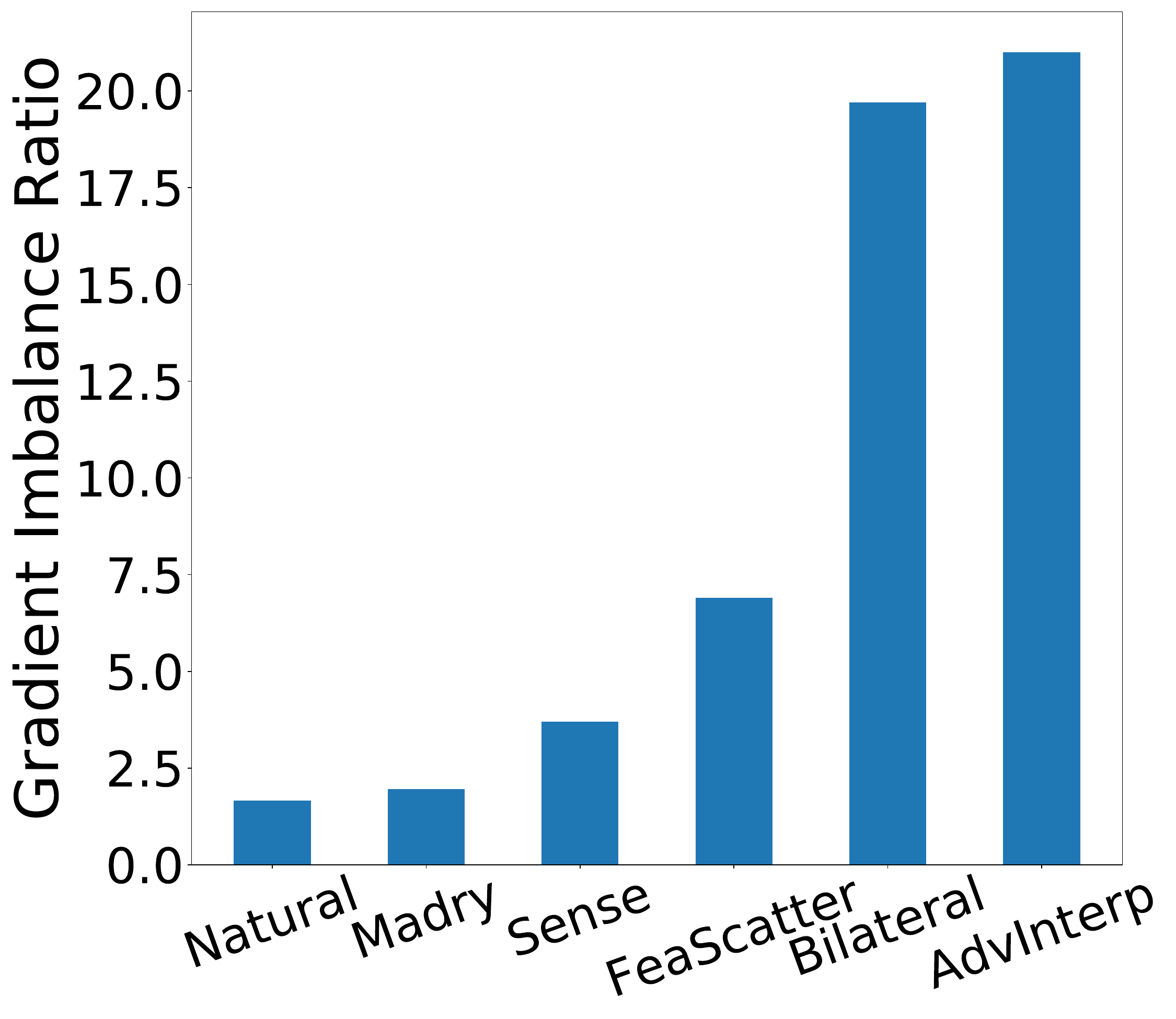}}%
    \hfill
    \subcaptionbox{ \label{fig:difference}}{\includegraphics[width=0.33\textwidth]{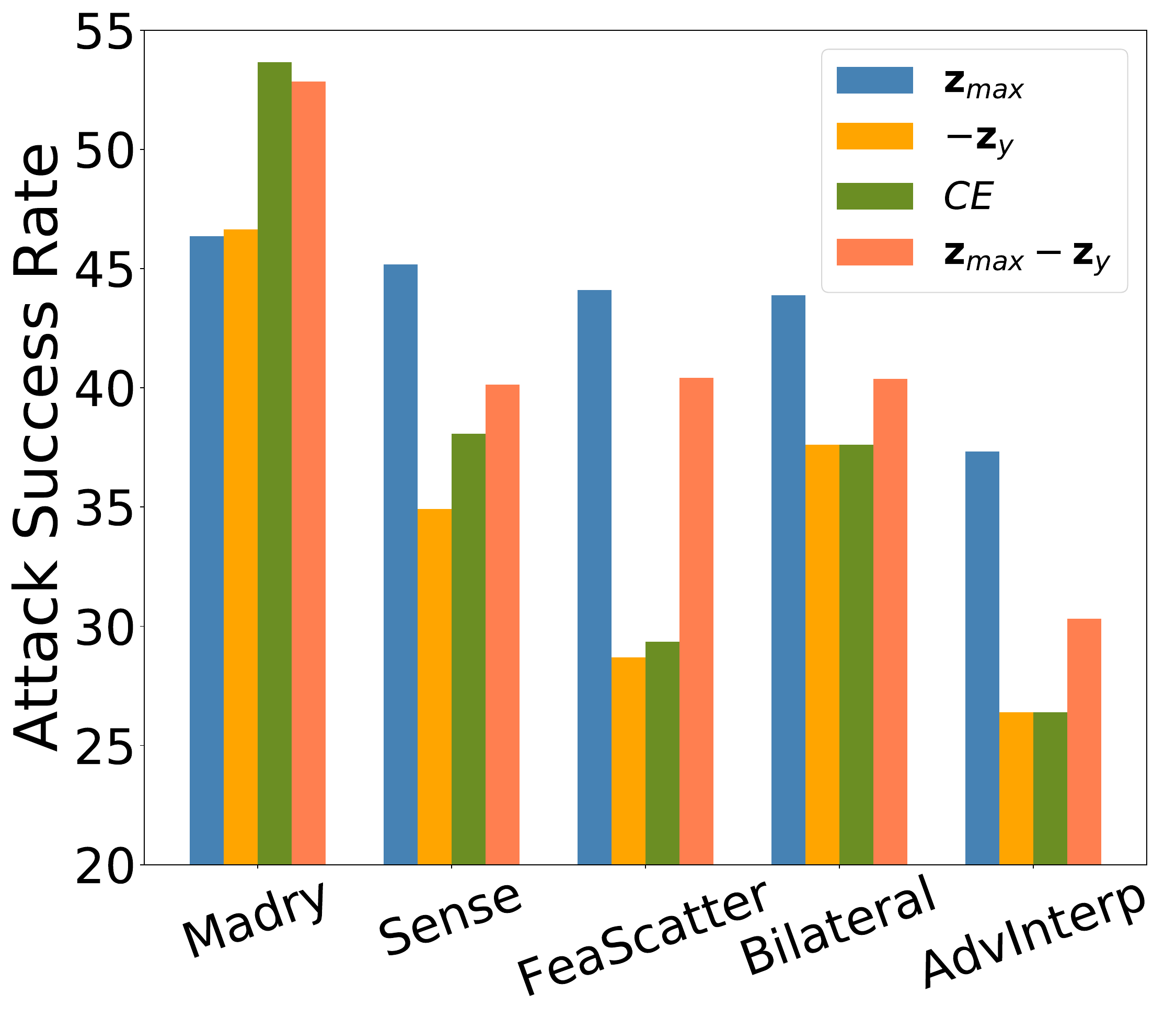}}%
    \hfill
    \subcaptionbox{ \label{fig:flat}}{\includegraphics[width=0.33\textwidth]{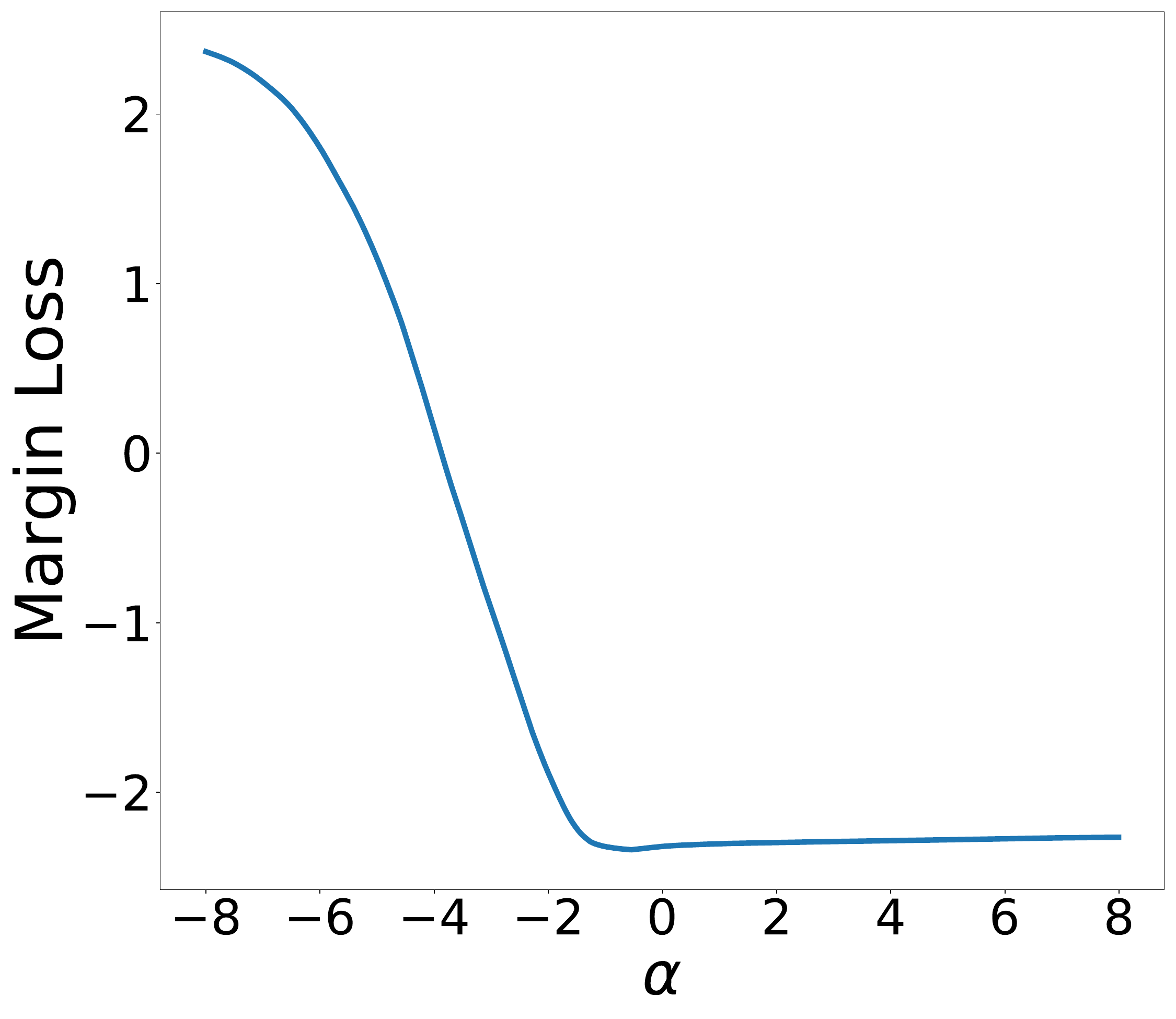}}
    \caption{(a): Gradient imbalance ratio of 5 models. (b): Attack success rate of PGD-20 with different losses.
    (c): The margin loss of the AdvInterp defense model on points $\bm{x}^* = \bm{x} + \alpha \cdot \sign(\nabla_{\bm{x}} (-\bm{z}_y))$, where $\bm{x}$ is a natural sample and $\sign(\nabla_{\bm{x}} (-\bm{z}_y))$ is the signed gradient of loss term $-\bm{z}_y$. All these experiments are conducted on test images of CIFAR-10.}
    \vskip -0.2in
\end{figure*}

Note that the GIR metric is not a general measure of imbalance. Rather, it is designed only for assessing gradient imbalance for \emph{adversarial robustness evaluation}. GIR focuses specifically on the imbalanced input dimensions and uses the $L_1$ norm to take into account the influence of these dimensions to model output. The ratio reflects how far away the imbalance towards one direction than the other.

% \vspace{0.05in}
\noindent\textbf{GIR of both Naturally- and Adversarial-trained DNNs.}
With the GIR metric, we next investigate 6 DNN models including a naturally-trained (Natural) model and 5 adversarially-trained models using standard adversarial training \citep{madry2017towards} (SAT), sensible adversarial training \citep{kim2020sensible} (Sense), feature scattering-based adversarial training \citep{zhang2019featurescatter} (FeaScatter), bilateral adversarial training \citep{wang2019bilateral} (Bilateral), and adversarial interpolation training \citep{zhang2020adversarial} (AdvInterp). We present these defense models here because they (except SAT) represent different levels of gradient imbalance (more results are provided in Fig. \ref{fig:all-mir}).
Natural, SAT, and Sense are WideResNet-34-10 models, while others are WideResNet-28-10 models. We train Natural and SAT following typical settings in \cite{madry2017towards}, while others use their officially released models.
We compute the GIR scores of the 6 models based on 1000 randomly selected test samples and show them in Fig. \ref{fig:mir}.
One quick observation is that some defense models have a much higher imbalance ratio than either naturally trained or SAT models.
This confirms that gradient imbalance does exist in DNNs, and some defenses tend to train the model to have highly imbalanced gradients.
We will show in Section \ref{sec:experiments} that this situation of imbalanced gradients may cause overestimated robustness when evaluated by the commonly used PGD attack.

\subsection{Imbalanced Gradients Reduce Attack Effectiveness}
When there are imbalanced gradients, the attack can be pushed by the dominant term to produce weak attacks, and the non-dominant term alone can lead to more successful attacks. To illustrate this, in Fig. \ref{fig:difference}, we show the success rates of PGD attack on the above 5 defense models (Natural has zero robustness against PGD) with different losses: CE loss, margin loss, and the two individual margin terms. We consider 20-step PGD (PGD-20) attacks with step size $\epsilon/4$ and $\epsilon=8/255$ on all CIFAR-10 test images.
One may expect the two margin terms to produce less effective attacks, as they only provide partial information about the margin loss. This is indeed the case for the low gradient imbalance model SAT. However, for highly imbalanced models Sense, FeaScatter, Bilateral, and AdvInterp, attacking the $\bm{z}_{max}$ term produces even more powerful attacks than attacking the margin loss.
This implies that the gradient of the margin loss is shifted by the dominant term (e.g., $-\bm{z}_y$ in this case) towards a less optimal direction, which inevitably causes less powerful attacks. Compared between attacking CE loss and attacking $-\bm{z}_{y}$, they achieve a very close performance on imbalanced models. This shows a stronger dominant effect of $-\bm{z}_{y}$ in CE loss ($\ell_{ce}(\bm{x}, y)=\log(\sum^{C}_{i=1} e^{\bm{z}_i}) - \bm{z}_{y}$). 
It is worth mentioning that, while both GIR and this individual term-based test can be used to check whether there are significantly imbalanced gradients in a defense model, GIR alone cannot predict the amount of overestimated robustness. Fig. \ref{fig:flat} shows an example of how the $-\bm{z}_y$ term leads the attack to a suboptimal direction: the margin loss is flat at the $\nabla_{\bm{x}} (-\bm{z}_y)$ direction, yet increases drastically at an opposite direction. In this example, the attack can actually succeed if it increases (rather than decreases) $\bm{z}_y$.

\begin{figure}%[!ht]
\vskip -0.1in
	\centering
	\begin{subfigure}{0.32\linewidth}
	    \centering
		\includegraphics[width=\textwidth]{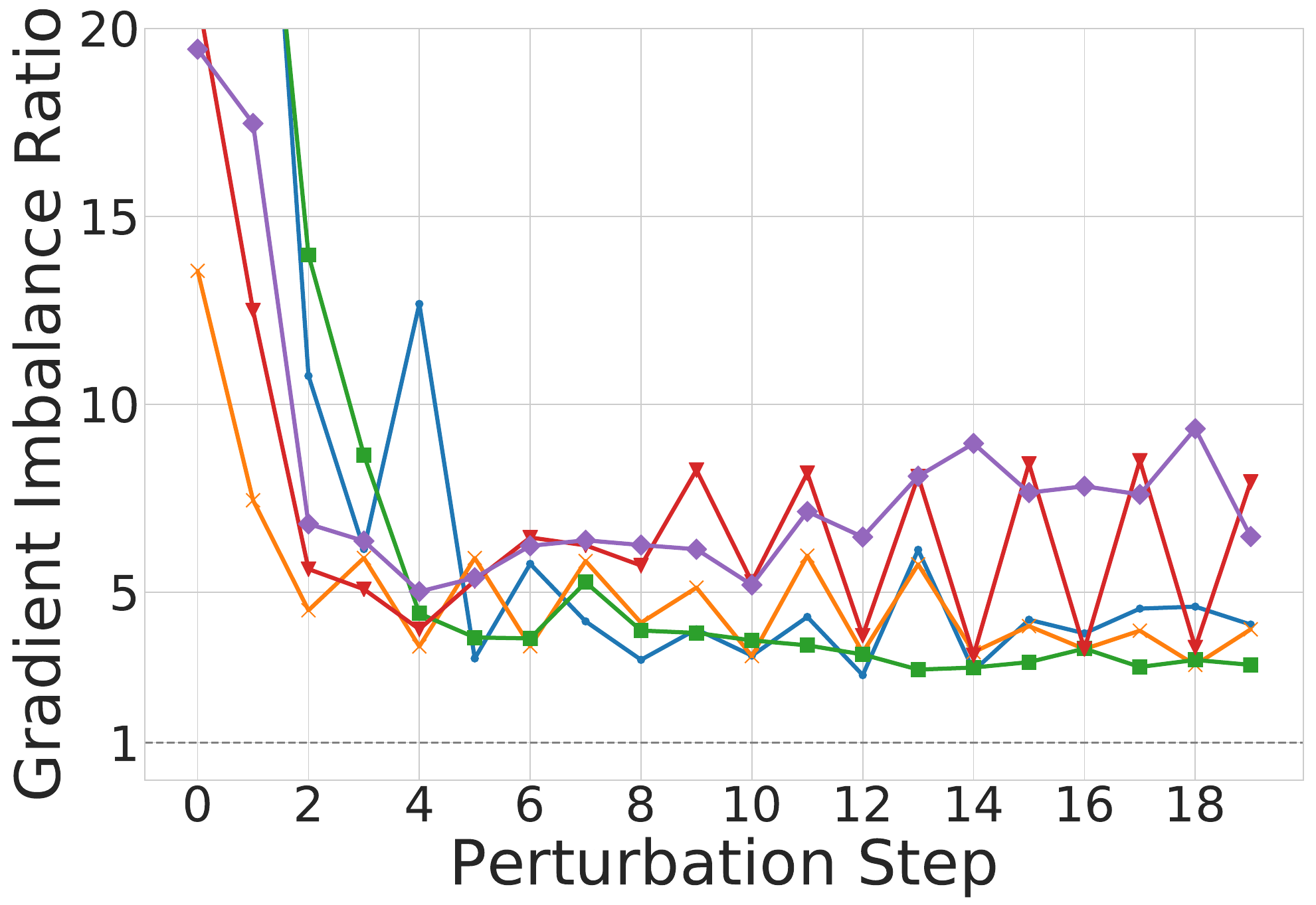}
		\caption{PGD-20 with loss $\zz_{max} - \zz_{y}$}
% 		\label{fig:toy-a}
	\end{subfigure}
    % \hspace{0.5cm}
	\begin{subfigure}{0.32\linewidth} 
	    \centering
		\includegraphics[width=\textwidth]{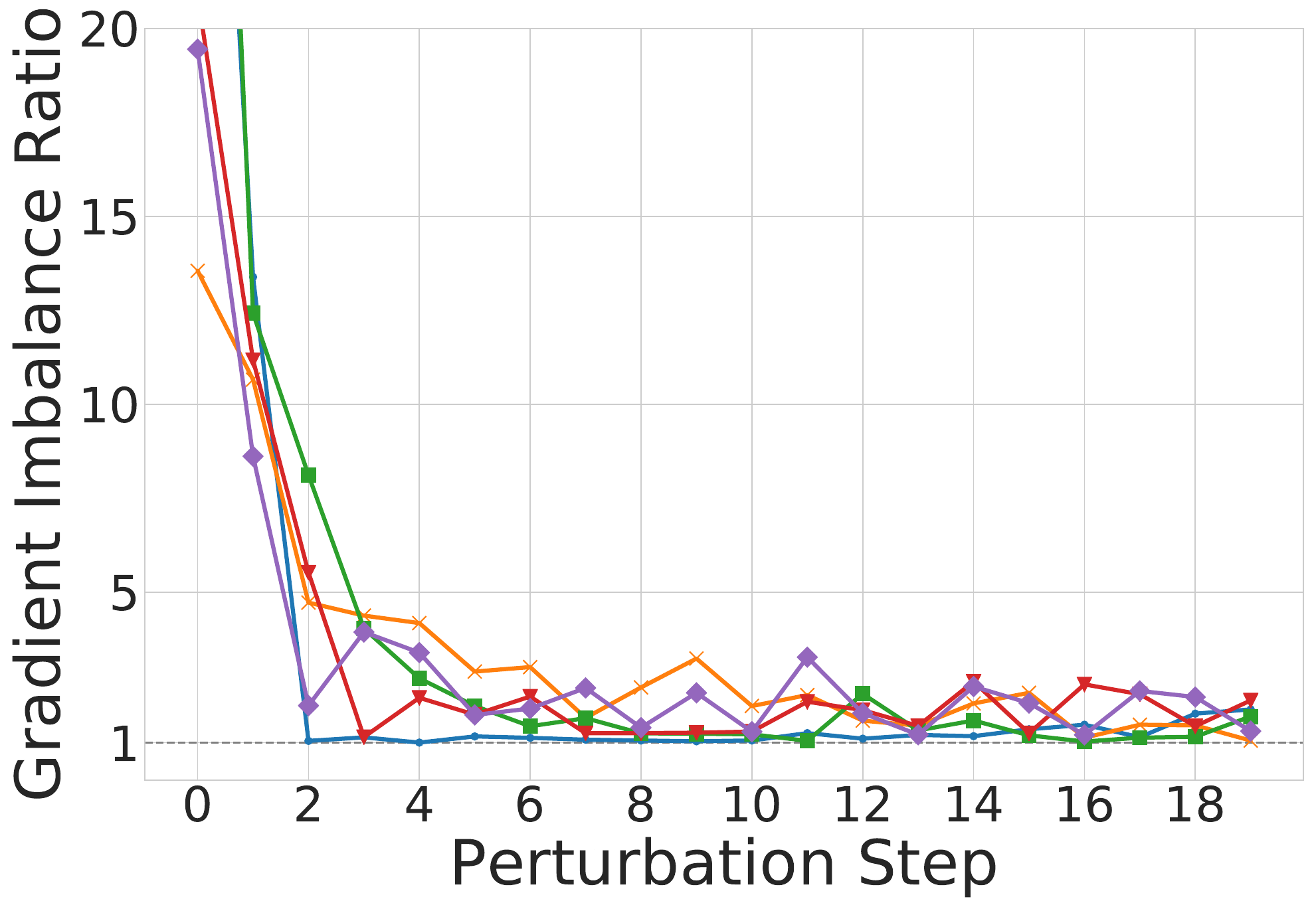}
		\caption{PGD-20 with loss $\zz_{max}$}
		\label{fig:3b}
	\end{subfigure}
	\begin{subfigure}{0.32\linewidth} 
	    \centering
		\includegraphics[width=\textwidth]{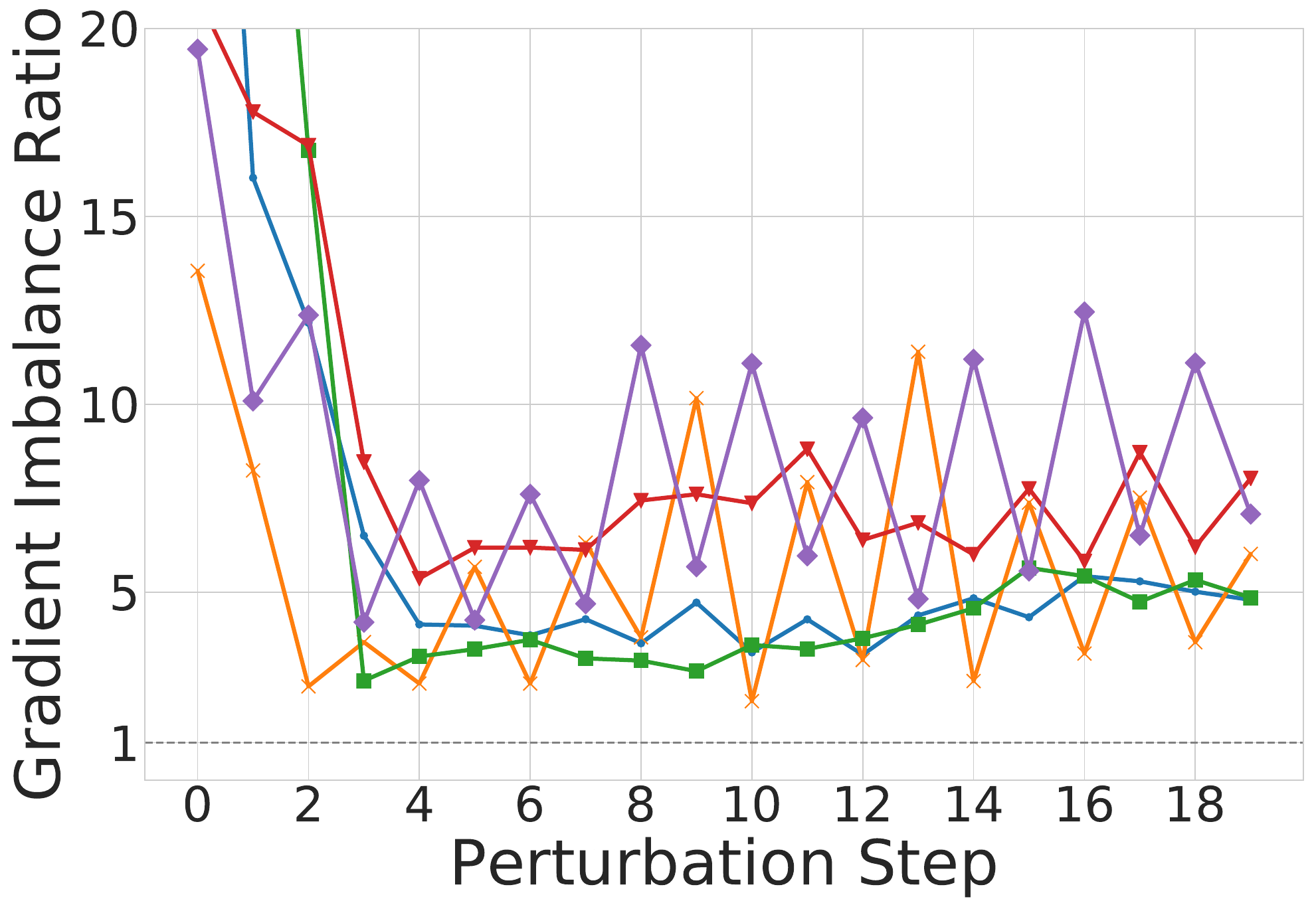}
		\caption{PGD-20 with loss $-\zz_{y}$}
% 		\label{fig:toy-b}
	\end{subfigure}
  \caption{Changes in gradient imbalance ratio when apply PGD-20 ($\epsilon=8/255$) attack with \textbf{a)} the margin loss, \textbf{b)} only the $\bm{z}_{max}$ term, or \textbf{c)} only the $-\bm{z}_{y}$ term, on the AdvInterp model for 5 CIFAR-10 test images. The imbalance ratio is effectively reduced by attacking a single $\bm{z}_{max}$ term. }
\label{fig:rebalance}
\vskip -0.2in
\end{figure}

\noindent\textbf{Gradients can be Balanced by Attacking Individual Loss Terms.}
Here, we show that, interestingly, imbalanced gradients can be balanced by attacking the non-dominant term. Consider the AdvInterp model tested above as an example, the dominant term is $-\bm{z}_{y}$. Fig. \ref{fig:rebalance} illustrates the GIR values of 5 randomly selected CIFAR-10 test images by attacking them using PGD-20 with different margin terms or the full margin loss.
As can be observed, for all three losses, the GIRs are effectively reduced after the first few steps. However, only the non-dominant term $\bm{z}_{max}$ manages to steadily reduce the imbalance ratio towards 1.
This indicates that optimizing the individual terms separately can help avoid the situation of imbalanced gradients and the attack can indeed benefit from more balanced gradients (see the higher success rate of $\bm{z}_{max}$ in Fig. \ref{fig:difference}).

\section{Proposed Margin Decomposition Attacks}
\noindent\textbf{Margin Decomposition Attack.} The above observations motivate us to exploit the individual terms in the margin loss so that the imbalanced gradients situation can be circumvented. Specifically, we propose Margin Decomposition (MD) attack that decomposes the attacking process with a margin loss into two stages: 1) attacking the two individual terms (e.g., $\bm{z}_{max}$ or $-\bm{z}_y$) alternately with restarts; then 2) attacking the full margin loss. Formally, our MD attack and its loss functions used in each stage are defined as follows:
\begin{align}
\label{eq:ld-pgd}
    \bm{x}_{k+1} &= \Pi_\epsilon \big(\bm{x}_k + \alpha\cdot\sign(\nabla_{\bm{x}} \ell^r_k(\bm{x}_k, y))\big),\\
    \ell^r_k(\bm{x}_k,y) &= 
    \begin{cases}
        \bm{z}_{max} & \text{if }k < K' \text{ and }r \bmod 2 = 0 \\
        -\bm{z}_{y} & \text{if }k < K' \text{ and } r \bmod 2 = 1 \\
        \bm{z}_{max} -\bm{z}_{y} & \text{if } K' \leq k \leq K,
    \end{cases}\nonumber
\end{align}
where, $\Pi$ is the projection operation that projects the perturbation back onto the $\epsilon$-ball around $\bm{x}$ if it goes beyond, $k \in \{1,\dots,K\}$ is the perturbation step, $K' \in [1, K]$ is the allocated step for the first stage (i.e., $\forall k \in [1, K')$), $r \in \{1,\dots,n\}$ is the $r$-th restart, $\bmod$ is the modulo operation for alternating optimization, and $\ell^r_k$ defines the loss function used at the $k$-th step and $r$-th restart. The loss function switches from the individual terms back to the full margin loss at step $K'$. The first stage exploits the two margin terms to rebalance the gradients, while the second stage ensures that the final objective (i.e., maximizing the classification error) is achieved. The complete algorithm of MD can be found in Algorithm \ref{alg:ld-pgd}.

Note that, not all defense models have the imbalanced gradients problem.
A model is susceptible to imbalanced gradients if there is a substantial difference between robustness evaluated by PGD attack and that by our MD attack.
In addition, to help escape the flat loss landscape observed in Fig. \ref{fig:flat}, we randomly initialize the perturbation at different restarts (line 6 in Algorithm \ref{alg:ld-pgd}), and use large initial perturbation size $\alpha=2\epsilon$ with cosine annealing for both stages (lines 8-12 in Algorithm \ref{alg:ld-pgd}).

\begin{algorithm}[!htb]
        \small
          \caption{Margin Decomposition Attack}
          \label{alg:ld-pgd}
        \begin{algorithmic}[1]
            \State {\bfseries Input:} clean sample $\bm{x}$, label $y$, model $f$, stage 1 steps $K'$, total steps $K$
            \State {\bfseries Output:}  adversarial example $\bm{x}_{adv}$
            \State {\bfseries Parameters:} Maximum perturbation $\epsilon$, step size $\alpha$, number of restarts $n$, first stage steps $K'$, total steps $K$
            \State $\bm{x}_{adv} \gets \bm{x}$
            \For{$ r \in \{1, ..., n\}$}
                \State $\bm{x}_0 \gets \bm{x} + uniform(-\epsilon,\epsilon)$ \Comment{uniform noise initialization}
                \For{$k \in \{1, ..., K\}$}
                \If{$k < K'$}
                    \State $\alpha \gets \epsilon\cdot \big(1+\cos(\frac{k-1}{K'}\pi)\big)$
                \ElsIf{$k \geq K'$}
                    \State $\alpha \gets \epsilon\cdot \big(1+\cos(\frac{k-K'}{K-K'}\pi)\big)$
                \EndIf
                    \State $\bm{x}_k \gets \Pi_\epsilon \big(\bm{x}_{k-1} + \alpha\cdot\sign(\nabla_{\bm{x}} \ell^r_k(\bm{x}_{k-1}, y))\big)$
                    \Comment{update $\bm{x}_k$ by Eqn. \eqref{eq:ld-pgd}}
                    % \bm{x}_{k+1} &= \Pi_\epsilon (\bm{x}_k + \alpha\cdot\sign(\nabla_{\bm{x}} \ell^r_k(\bm{x}_k, y)))
                    \If{$\ell(\bm{x}_{adv}) < \ell(\bm{x}_k)$}
                        \State $\bm{x}_{adv} \gets \bm{x}_k$
                    \EndIf
                \EndFor
            \EndFor
            \State $\bm{x}_{adv} = \Pi_{[0,1]}\big(\bm{x}_{adv}\big)$ \Comment{final clipping}
            \State {\bfseries return} $\bm{x}_{adv}$
        \end{algorithmic}
        \end{algorithm}

\noindent\textbf{Multi-Targeted MD Attack.}
We also propose a multi-targeted version of our MD attack and call it MD-MT. The loss terms used by MD-MT at different attacking stages are defined as follows:
\begin{align}
  \label{eq:ldmt}
  \bm{x}_{k+1} &= \Pi_\epsilon \big(\bm{x}_k + \alpha\cdot\sign(\nabla_{\bm{x}} \ell^r_k(\bm{x}_k, t))\big),\\
  \ell^r_k(\bm{x}_k,y) &= 
  \begin{cases}
      \bm{z}_{t} & \text{if }k < K' \text{ and }r \bmod 2 = 0 \\
      -\bm{z}_{y} & \text{if }k < K' \text{ and } r \bmod 2 = 1 \\
      \bm{z}_{t} -\bm{z}_{y} & \text{if } K' \leq k \leq K,
  \end{cases}\nonumber
\end{align}
where, $\bm{z}_{t}$ is the logits of a target class $t \neq y$. Other parameters are the same as in Eqn. \eqref{eq:ld-pgd}.
% and $K' \in [1, K]$ is the allocated step for the first stage (i.e., $\forall k \in [1, K')$).
% which will be switched to a different class at each restart. 
Like the MT attack, MD-MT will attack each possible target class one at a time, then select the strongest adversarial example at the end. That is, the target class $t \neq y$ will be switched to a different target class at each restart.
The complete algorithm MD-MT can be found in Algorithm \ref{alg:MD-MT}.
% and an ablation study can be found in Appendix \ref{sec:ablation}.

\begin{algorithm}[!htb]
  \caption{MultiTargeted Margin Decomposition Attack}
  \label{alg:MD-MT}
\begin{algorithmic}[1]
    \State {\bfseries Input:} clean sample $\bm{x}$, class label $y$, class set $\mathcal{T}$, model $f$
    \State {\bfseries Output:}  adversarial example $\bm{x}_{adv}$
    \State {\bfseries Parameters:} Maximum perturbation $\epsilon$, step size $\alpha$, number of restarts $n$, first stage steps $K'$, total steps $K$
    
    \State $n_\textrm{r} \gets \left\lfloor n / \lvert\mathcal{T}\rvert \right\rfloor$, $\bm{x}_{adv} \gets \bm{x}$
    \For{$ r \in \{1, ..., n_\textrm{r}\}$}
        % \State $t \gets \text{r mode } |\mathcal{T}|$
        \For{$ t \in \mathcal{T}$}
            \State $\bm{x}_0 \gets \bm{x} + uniform(-\epsilon,\epsilon)$ \Comment{uniform noise initialization}
            \For{$k \in \{1, ..., K\}$}
            \If{$k < K'$}
                \State $\alpha \gets \epsilon\cdot \big(1+\cos(\frac{k-1}{K'}\pi)\big)$
            \ElsIf{$k \geq K'$}
                \State $\alpha \gets \epsilon\cdot \big(1+\cos(\frac{k-K'}{K-K'}\pi)\big)$
            \EndIf
            \State Update $\bm{x}_k$ by Eqn. \eqref{eq:ldmt}
            \If{$\ell(\bm{x}_{adv}) < \ell(\bm{x}_k)$}
                \State $\bm{x}_{adv} \gets \bm{x}_k$
            \EndIf
            \EndFor
        \EndFor
    \EndFor
    \State $\bm{x}_{adv} = \Pi_{[0,1]}\big(\bm{x}_{adv}\big)$ \Comment{final clipping}
    \State {\bfseries return} $\bm{x}_{adv}$

\end{algorithmic}
\end{algorithm}

\noindent\textbf{MD Ensemble Attack.}
Following AutoAttack \citep{croce2020reliable}, here we also propose an ensemble attack to fully exploit the strengths of both existing attacks and the gradient exploration of our MD attacks. 
% The principle is to replace the two in AutoAttack with our MD attacks. 
The AutoAttak ensemble consists of 4 attacks: 1) APGD$_{CE}$, which is the Auto-PGD with the cross entropy loss; 2) DLR, which is the Auto-PGD with the Difference of Logits Ratio (DLR) loss; 3) Fast Adaptive Boundary Attack (FAB) \citep{croce2019minimally}; and 4) the black-box Square Attack \citep{andriushchenko2019square}. The MultiTargeted version of both APGD$_{CE}$ and DLR are used in the latest version of AutoAttack.
We first replace the MultiTargeted DLR (DLR-MT), the strongest attack in the AutoAttack ensemble with our MD-MT attack. We then replace the Square attack with our MD attack as we focus on white-box robustness and gradient issues.
This gives us the MD Ensemble (MDE) of 4 attacks including 1) \textbf{MD}, 2) \textbf{MD-MT}, 3) \textbf{APGD$_{CE}$} and 4) \textbf{FAB}. We did not replace the APGD$_{CE}$ attack as we find it is better to have a cross entropy loss based attack in the ensemble.

\noindent\textbf{Initialization Perspective Interpretation of MD Attacks.} Previous works have shown that random or logits diversified initialization are crucial for generating strong adversarial attacks \citep{madry2017towards,tashiro2020ods}. Compared to random or logits diversified initialization, our MD attacks can be interpreted as a type of \emph{adversarial initialization}, i.e., initialization at the adversarial sub-directions defined by the two margin terms.  Rather than a single step of initialization, our MD attacks iteratively explore the optimal starting point during the first attacking stage.

\noindent\textbf{Extending MD to Complex Losses with More Than Two Terms}. Our MD strategy is not restricted to the margin loss, it is also suitable for other margin-based losses like CW and DLR attacks. For more complex loss functions with more than two terms, one can group the individual terms into two conflicting groups that could produce opposite gradient directions or investigate the two most conflicting terms (in the form of ‘A - B’). In this way, a complex loss could be reformulated in a margin form where our margin decomposition strategy can be easily applied.

\section{Experiments}\label{sec:experiments}
\noindent\textbf{Defense Models.} We apply our MD attacks to evaluate the robustness of 24 defense models proposed since 2018. Here, we focus on adversarial training models, which are arguably the strongest defense approach to date \citep{athalye2018obfuscated,croce2020reliable}.
The selected defense models are as follows. 

The Standard Adversarial Training (SAT) \citep{madry2017towards} trains models on adversarial examples generated by PGD attack.
Dynamic adversarial training (Dynamic) \citep{wang2019convergence} trains on adversarial examples with gradually increased convergence quality.
Max-Margin Adversarial training (MMA) \citep{ding2018mma} trains on adversarial examples with increasing margins (e.g., the perturbation bound $\epsilon$). For MMA, we evaluate the released ``MMA-32'' model. 
Jacobian Adversarially Regularized Networks (JARN) adversarially regularize the Jacobian matrices and can be combined with 1-step adversarial training (JARN-AT1) to gain additional robustness \citep{chan2020jacobian}. For JARN, we only evaluate the JARN-AT1 as its none adversarial training version has been completely broken in \citep{croce2020reliable}. We implement JARN-AT1 on the basis of their released implementation of JARN.
Sensible adversarial training (Sense) \citep{kim2020sensible} trains on loss-sensible adversarial examples (perturbation stops when the loss exceeds a certain threshold). 
Bilateral adversarial training (Bilateral) \citep{wang2019bilateral} trains on PGD adversarial examples with adversarially perturbed labels. For Bilateral, we mainly evaluate its released strongest model ``R-MOSA-LA-8''. Adversarial Interpolation (Adv-Interp) training \citep{zhang2020adversarial} trains on adversarial examples generated under an adversarial interpolation scheme with adversarial labels.
Feature Scattering-based (FeaScatter) adversarial training \citep{zhang2019featurescatter} crafts adversarial examples using latent space feature scattering, then trains on these examples with label smoothing.
Adversarial Training with Hypersphere Embedding (AT-HE) \citep{pang2020boosting} regularizing the features onto compact manifolds.
Adversarial Training with Pre-Training (AT-PT) \citep{hendrycks2019using} uses pre-training to improve model robustness.
TRADES \citep{zhang2019trades} replaces the CE loss of SAT with the KL divergence for a better trade-off between robustness and natural accuracy. Based on TRADES, RST \citep{carmon2019unlabel} and UAT \citep{stanforth2019labels} improve robustness by training with $10\times$ more unlabeled data. Misclassification Aware adveRsarial Training (MART) \citep{wang2020improving} further improves the above three methods with a misclassification aware loss function. 
Adversarial Training with Early Stopping (AT-ES) \citep{rice2020overfitting} improves SAT by using early stopping to avoid robust overfitting. 
Adversarial Weight Perturbation (AWP) \citep{wu2020adversarial} proposes a double perturbation mechanism to explicitly regularize the flatness of the weight loss landscape.
Robust WideResNet (R-WRN) \citep{huang2021exploring} explores the most robust configurations of WideResNet \citep{Zagoruyko16wideresnet} and trains the model following the same procedure as RST. This robust configuration can bring additional stability to the model and improve robustness. The 4 defenses we consider for the ImageNet dataset are as follows.
1) \cite{engstrom2019adversarial} showed that robust representations obtained via adversarial training on ImageNet are approximately invertible. 2) Fast adversarial training (FastAT) \citep{wong2020fast} has also been found to be crucial for efficient adversarial training on ImageNet. 3) \cite{salman2020adversarially} trained adversarially robust models on ImageNet to demonstrate their transferability to other tasks. 4) A recent work by \cite{debenedetti2022light} explored the adversarial training hyperparameters for vision transformers and trained robust XCiT models on ImageNet.

The architectures of the defense from CIFAR-10 models are all WideResNet variants \citep{Zagoruyko16wideresnet}. We also evaluated other architectures, including VGG-19 \citep{simonyan2014very}, DenseNet-121 \citep{huang2017densely}, and DARST \citep{liu2018darts} trained with SAT \citep{madry2017towards}. The configuration for these models follows the same setting as in \cite{huang2021exploring}. 
For each defense model, we either download their shared models or retrain the models using the official implementations, unless explicitly stated. For ImageNet, we consider the adversarially pre-trained ResNet-50 and vision transformer XCiT-S models.
% Further details about the models can be found in Appendix \ref{sec:defenses}. 
The defense models were trained against maximum perturbation $\epsilon = 8/255$ on CIFAR-10 and $\epsilon=4/255$ on ImageNet.
We apply the current state-of-the-art and our MD attacks to evaluate the robustness of these models in a white-box setting.

\noindent\textbf{Baseline Attacks and Settings.}
Following the current literature, we consider 4 existing untargeted attacks: 1) the $L_{\infty}$ version of CW attack \citep{madry2017towards,wang2019convergence}, 2) Projected Gradient Descent (PGD) \citep{madry2017towards}, 3)  Output Diversified Initialization (ODI) \citep{tashiro2020ods}, 4) fast adaptive boundary attack (FAB) \citep{croce2020minimally}, 5) Auto-PGD with Difference of Logits Ratio (DLR) \citep{croce2020reliable}.
We consider 3 multi-target attacks, 1) Multi-Targeted (MT) attack \citep{gowal2019multitargeted}, 2) Multi-Targeted FAB (FAB-MT), 3) Multi-Targeted DLR (DLR-MT). 
% Note that DLR is the strongest standalone attack in the AutoAttack ensemble.
The evaluation was conducted under the same maximum perturbation $\epsilon = 8/255$ and $\epsilon = 4/255$ for CIFAR-10 and ImageNet respectively. For all untargeted attacks, we use the same total perturbation steps $K=100$. For DLR and FAB, we use the official implementation and parameter setting. For PGD and ODI, we use 5 restarts and step size set to $\alpha=0.8/255$. 
For both stages of our MD attack, we use a large initial step size $\alpha=2\epsilon$, then gradually decrease it to 0 via cosine annealing. The number of steps is set to $K'=20$ for the first stage (i.e., $K-K'=80$ for the second stage). 
% The second stage of our MD attack uses a typical PGD setting with step size $\alpha=2\epsilon$
For all multi-target attacks, we use the same total perturbation steps $K=100$ for each class. For CIFAR-10, this means a total of 900 steps for each image
 (as there are only 9 possible target classes).
For DLR-MT and FAB-MT, we use the official implementation and parameter setting.
For MT attack, we use 5 restarts and step size set to $\alpha=0.8/255$. 
For our MD-MT attack, we use the same parameter setting as the untargeted MD when attacking each target class. 
For a fair comparison, the total number of perturbation steps is set to be the same for all attacks.
We also compare our MD Ensemble (MDE) with the AutoAttack ensemble. 
% For MDE, we replace DLR-MT and Square attack in AutoAttack to MD and MD-MT attack.
Adversarial robustness is measured by the model's accuracy on adversarial examples crafted by the attack on CIFAR-10 and ImageNet test images.
We excluded FAB from the untargeted evaluation on ImageNet and all ImageNet models in targeted and ensemble evaluations due to the attack's efficiency.

\begin{table}[ht!]
\centering
\caption{\textbf{Untargeted Evaluation}. Robustness ($\%$) of 24 defense models on CIFAR-10 and 4 defense models on ImageNet, all evaluated by untargeted attacks. The `diff' column shows the robustness decrease by our MD attack compared to the \emph{best} baseline attack (i.e., best baseline - MD). The best results are \textbf{boldfaced}.}
\label{tbl:comprehensive}
\begin{adjustbox}{width=\linewidth}
\begin{tabular}{l|c|c|ccccccc}
\hline
\multicolumn{1}{c|}{Dataset} & \multicolumn{1}{c|}{Defense (\textit{ranked by MD robustness})} & \multicolumn{1}{c|}{Clean} & CW & PGD & ODI & FAB & DLR & \textbf{MD} & \textbf{Diff} \\ \hline
\multicolumn{1}{l|}{\multirow{20}{*}{CIFAR10}} & R-WRN \citep{huang2021exploring} & 91.23 & 64.28 & 64.98 & 64.76 & 64.49 & 64.38 & \textbf{62.91} & \textbf{-1.37}       \\
\multicolumn{1}{l|}{} & AWP \citep{wu2020adversarial} & 88.25 & 63.64 & 63.45 & 62.14 & 60.74 & 60.73 & \textbf{60.19} & \textbf{-0.54} \\
\multicolumn{1}{l|}{} & RST \citep{carmon2019unlabel} & 89.69 & 62.25 & 61.99 & 61.89 & 60.92 & 60.88 & \textbf{59.77} & \textbf{-1.11} \\
\multicolumn{1}{l|}{} & UAT \citep{stanforth2019labels} & 86.46 & 61.19 & 60.96 & 60.17 & 60.06 & 62.97 & \textbf{58.44} & \textbf{-1.62} \\
\multicolumn{1}{l|}{} & AT-PT \citep{hendrycks2019using} & 87.11 & 57.57 & 57.53 & 56.84 & 55.57 & 57.25 & \textbf{55.22} & \textbf{-0.35} \\
\multicolumn{1}{l|}{} & AT-ES \citep{rice2020overfitting} & 85.52 & 55.80 & 55.80 & 54.50 & 53.14 & 54.67 & \textbf{52.94} & \textbf{-0.20} \\
\multicolumn{1}{l|}{} & TRADES \citep{zhang2019trades} & 84.92 & 55.07 & 54.98 & 54.83 & 53.50 & 53.66 & \textbf{52.74} & \textbf{-0.76} \\
\multicolumn{1}{l|}{} & AT-HE \citep{pang2020boosting} & 81.43 & 53.05 & 52.87 & 52.89 & 52.20 & 54.62 & \textbf{51.85} & \textbf{-0.35} \\
\multicolumn{1}{l|}{} & MART \citep{wang2020improving} & 83.62 & 56.09 & 55.84 & 53.43 & 51.80 & 52.90 & \textbf{51.50} & \textbf{-0.30} \\
\multicolumn{1}{l|}{} & SAT-DenseNet121 \citep{huang2021exploring} & 86.07 & 52.75 & 52.70 & 53.01 & 51.39 & 53.06 & \textbf{50.83} & \textbf{-0.56} \\
\multicolumn{1}{l|}{} & SAT-DARTS \citep{huang2021exploring} & 86.76 & 49.16 & 46.35 & 48.64 & 46.65 & 47.38 & \textbf{45.78} & \textbf{-0.57} \\
\multicolumn{1}{l|}{} & Adv-Interp \citep{zhang2020adversarial} & 90.25 & 73.12 & 73.05 & 50.85 & \textbf{43.96} & 52.87 & 45.07 & 1.11 \\
\multicolumn{1}{l|}{} & MMA \citep{ding2018mma} & 84.36 & 51.34 & 51.40 & 46.90 & 48.97 & 51.20 & \textbf{44.94} & \textbf{-1.96} \\
\multicolumn{1}{l|}{} & SAT \citep{madry2017towards} & 87.25 & 51.49 & 45.32 & 47.23 & 45.94 & 46.85 & \textbf{44.70} & \textbf{-0.62} \\
\multicolumn{1}{l|}{} & Dynamic \citep{wang2019convergence} & 84.36 & 51.34 & 51.40 & 44.30 & 46.90 & 51.20 & \textbf{44.94} & \textbf{-0.48} \\
\multicolumn{1}{l|}{} & SAT-VGG19 \citep{huang2021exploring} & 77.06 & 45.51 & 46.06 & 44.86 & 43.88 & 46.08 & \textbf{43.56} & \textbf{-0.32} \\
\multicolumn{1}{l|}{} & FeaScatter \citep{zhang2019featurescatter} & 89.98 & 69.56 & 69.35 & 44.89 & 43.78 & 51.82 & \textbf{42.16} & \textbf{-1.62} \\
\multicolumn{1}{l|}{} & Sense \citep{kim2020sensible} & 91.51 & 60.68 & 60.61 & 46.32 & 42.24 & 50.54 & \textbf{39.91} & \textbf{-2.33}\\
\multicolumn{1}{l|}{} & Bilateral \citep{wang2019bilateral} & 90.73 & 61.50 & 61.20 & 44.73 & 43.79 & 46.36 & \textbf{39.39} & \textbf{-4.11} \\
\multicolumn{1}{l|}{} & JARN-AT1 \citep{chan2020jacobian} & 83.86 & 50.15 & 19.36 & 19.73 & 20.14 & 20.57 & \textbf{17.32} & \textbf{-2.04} \\ \hline
\multicolumn{1}{l|}{\multirow{4}{*}{ImageNet}} & XCiT-S \citep{debenedetti2022light} & 72.53 & 55.55 & 42.84 & 41.63 & - & 44.48 & \textbf{40.26} & \textbf{-1.37} \\
\multicolumn{1}{l|}{} & ResNet-50 \citep{salman2020adversarially} & 63.87 & 51.14 & 38.60 & 36.41 & - & 37.92 & \textbf{35.31} & \textbf{-1.09} \\
\multicolumn{1}{l|}{} & ResNet-50 \citep{engstrom2019adversarial} & 62.40 & 49.16 & 32.46 & 31.23 & - & 33.00 & \textbf{29.76} & \textbf{-1.47} \\
\multicolumn{1}{l|}{} & FastAT \citep{wong2020fast} & 53.82 & 44.55 & 27.28 & 26.31 & - & 27.48 & \textbf{25.34} & \textbf{-0.98} \\ \hline
\end{tabular}
\end{adjustbox}
\end{table}

\begin{table}[ht!]
\centering
\caption{\textbf{Multi-targeted Evaluation}. Robustness ($\%$) of 20 defense models on CIFAR-10 evaluated by different multi-targeted attacks. The `diff' column shows the robustness decrease by our MD-MT attack compared to the \emph{best} baseline attack (i.e., best baseline - MD-MT). The best results are \textbf{boldfaced}.}
\label{tbl:comprehensive_mt}
\begin{adjustbox}{width=\linewidth}
\begin{tabular}{l|c|ccccc}
\hline
\multicolumn{1}{c|}{Defense (\textit{ranked by MD-MT robustness})} & Clean & MT & FAB-MT & DLR-MT & MD-MT  & Diff  \\ \hline
R-WRN \citep{huang2021exploring} & 91.23 & 62.64 & 63.18 & \textbf{62.55} & 62.57 & 0.02   \\
AWP \citep{wu2020adversarial} & 88.25 & 60.12 & 60.52 & \textbf{60.05}  & 60.07 & 0.02 \\
RST \citep{carmon2019unlabel} & 89.69 & 59.74 & 60.20 & 59.58  & \textbf{59.55} & \textbf{-0.03} \\
UAT \citep{stanforth2019labels} & 86.46 & 56.50 & 59.98 & \textbf{56.16} & 56.19 & 0.03 \\
AT-PT \citep{hendrycks2019using} & 87.11 & 55.04 & 55.28 & 54.88 & \textbf{54.87} & \textbf{-0.01} \\
TRADES \citep{zhang2019trades} & 84.92 & 52.61 & 53.06 & 52.53  & \textbf{52.51} & \textbf{-0.02} \\
AT-ES \citep{rice2020overfitting} & 85.52 & 52.42 & 52.75 & 52.35  & \textbf{52.31} & \textbf{-0.04} \\
AT-HE \citep{pang2020boosting} & 81.43 & 51.17 & 51.53 & 51.10  & \textbf{51.04} & \textbf{-0.06} \\
MART \citep{wang2020improving} & 83.62 & 51.00 & 51.39 & 50.95  & \textbf{50.91} & \textbf{-0.04} \\
SAT-DenseNet121 \citep{huang2021exploring} & 86.07 & 50.24 & 50.64 & \textbf{50.12} & 50.14 & 0.02 \\
SAT-DARTS \citep{huang2021exploring} & 86.76 & 45.39 & 45.97 & \textbf{45.09} & 45.11 & 0.02 \\
SAT \citep{madry2017towards} & 87.25 & 44.67 & 45.29 & 44.52 & \textbf{44.49} & \textbf{-0.03}  \\
Dynamic \citep{wang2019convergence} & 84.48 & 43.01 & 43.40 & 42.93  & \textbf{42.91} & \textbf{-0.02} \\
SAT-VGG19 \citep{huang2021exploring} & 77.06 & 42.67 & 43.30 & 42.62 & \textbf{42.60} & \textbf{-0.02} \\
MMA \citep{ding2018mma} & 84.36 & 42.74 & 42.66 & 41.72  & \textbf{41.45} & \textbf{-0.27} \\
ADVInterp \citep{zhang2020adversarial} & 90.25 & 66.30 & 39.10 & \textbf{37.53}  & 37.54 & 0.01 \\
Bilateral \citep{wang2019bilateral} & 90.73 & 57.51 & 38.36 & 38.55  & \textbf{37.03} & \textbf{-1.33} \\
FeaScatter \citep{zhang2019featurescatter} & 89.98 & 43.37 & 38.54 & 37.29  & \textbf{36.72} & \textbf{-0.57} \\
Sense \citep{kim2020sensible} & 91.51 & 48.40 & 35.50 & 35.94  & \textbf{34.87} & \textbf{-0.63} \\
JARN-AT1 \citep{chan2020jacobian} & 83.86 & 17.10 & 
17.49 & \textbf{16.58} & 16.63 & 0.05 \\ \hline
\end{tabular}
\end{adjustbox}
\end{table}

\begin{table}[ht!]
\centering
\caption{\textbf{Ensemble Evaluation.} Robustness ($\%$) of 20 defense models evaluated by ensemble attacks. The `diff' column shows the robustness decrease by our MDE attack compared to the \emph{best} baseline attack (i.e., best baseline - MDE). The best results are \textbf{boldfaced}.}
\label{tbl:comprehensive_mde}
\begin{tabular}{l|c|ccc}
\hline
\multicolumn{1}{c|}{Defense (ranked)}& Clean & AutoAttack & MDE   & Diff  \\ \hline
R-WRN \citep{huang2021exploring} & 91.23 & 62.54 & 62.54 & 0.00 \\
AWP \citep{wu2020adversarial} & 88.25 & 60.05      & \textbf{60.00} & \textbf{-0.05} \\
RST \citep{carmon2019unlabel} & 89.69 & 59.56      & \textbf{59.53} & \textbf{-0.03} \\
UAT \citep{stanforth2019labels} & 86.46 & \textbf{56.11}      & 56.16 & 0.05 \\
AT-PT \citep{hendrycks2019using} & 87.11 & 54.91      & \textbf{54.86} & \textbf{-0.05} \\
TRADES \citep{zhang2019trades} & 84.92 & 52.54      & \textbf{52.51} & \textbf{-0.03} \\
AT-ES \citep{rice2020overfitting} & 85.52 & 52.34      & \textbf{52.30} & \textbf{-0.04} \\
AT-HE \citep{pang2020boosting} & 81.43 & 51.09      & \textbf{51.06} & \textbf{-0.03} \\
SAT-DenseNet121 \citep{huang2021exploring} & 86.07 & 50.11 & 50.09 & \textbf{-0.02} \\
SAT-DARTS \citep{huang2021exploring} & 86.76  & \textbf{45.00}  & 45.01 & 0.01 \\
MART \citep{wang2020improving} & 83.62 & 50.94      & \textbf{50.89} & \textbf{-0.05} \\
SAT \citep{madry2017towards} & 87.25 & 44.45 & 44.45 & 0.00 \\
Dynamic \citep{wang2019convergence} & 84.48 & 42.90      & \textbf{42.89} & \textbf{-0.01} \\
SAT-VGG19 \citep{huang2021exploring} & 77.06 & 42.61 & \textbf{42.57} & \textbf{-0.04} \\
MMA \citep{ding2018mma}      & 84.36 & 41.51      & \textbf{41.34} & \textbf{-0.17} \\
ADVInterp \citep{zhang2020adversarial} & 90.25 & 36.46  & \textbf{36.55} & \textbf{-0.09} \\
Bilateral \citep{wang2019bilateral} & 90.73 & 36.61      & \textbf{36.48} & \textbf{-0.13} \\
FeaScatter \citep{zhang2019featurescatter} & 89.98 & 36.65      & \textbf{36.25} & \textbf{-0.40} \\
Sense \citep{kim2020sensible} & 91.51 & 34.19      & \textbf{33.84} & \textbf{-0.35} \\
JARN-AT1 \citep{chan2020jacobian} & 83.80 & 16.55 & \textbf{16.51} & \textbf{-0.04} \\ \hline
\end{tabular}
\end{table}

\subsection{Robustness Evaluation Results}
Table \ref{tbl:comprehensive} reports the untargeted evaluation result, where R-WRN, AWP, and RST are the top 3 best defenses.
The `Diff' column shows that there are 11 defense models demonstrating more than 1\% robustness drop against our MD attack compared to the best baseline attack. This implies that these models are susceptible to imbalanced gradients. Out of the 11 models, 4 of them are ranked at the very bottom (worst robustness on CIFAR-10) of the list with a robustness that is much lower than SAT.
In most cases, our MD attack is able to decrease the PGD or DLR robustness by more than 2\% even on the top 5 defense models. 
Compared to the best baseline attack, our MD attack can reduce its evaluated robustness by more than 2\% on Sense, Bilateral, FeaScatter, and JARN-AT1. It also shows more than 1\% improvement on top-ranking defenses, including R-WRN, RST, and UAT. These results demonstrate the importance of addressing the imbalanced gradients problem in robustness evaluation.
Circumventing imbalanced gradients via margin decomposition and exploration makes our MD the best overall standalone attack for robustness evaluation. 
In terms of evaluating ImageNet defenses, our MD attack alone can outperform the AA ensemble attack on XCiT-S and FastAT, compared to the robustness evaluated and reported on RobustBench leaderboard \citep{croce2021robustbench}. For instance, AA evaluates XCiT-S to be of 41.78\% robustness, while our MD evaluates it to be of 40.26\% robustness. This highlights the significance of the imbalanced gradients problem on the large-scale dataset and the advantage of our proposed technique in evaluating vision transformers.

The multi-targeted evaluation results are presented in Table \ref{tbl:comprehensive_mt}, where it shows that our MD-MT attack is the strongest attack on average. By comparing Table \ref{tbl:comprehensive} with Table \ref{tbl:comprehensive_mt}, we find that the multi-targeted robustness is lower than the untargeted robustness for all defenses. This indicates that multi-targeted evaluation is more accurate than untargeted evaluation. Overall, our MD-MT attack demonstrates the most reliable robustness on 13/20 defense models, while DLR-MT is slightly better on the other 7 models. The improvement of DLR-MT over DLR indicates that exploring different targets can help circumvent imbalanced gradients to some extent. We will conduct a detailed analysis of different attack techniques against imbalanced gradients in Section \ref{sec:5.3}.

We further compare our MDE attack with the AutoAttack in Table \ref{tbl:comprehensive_mde}. Compared to multi-targeted evaluation, the use of ensemble attacks produces the most accurate evaluation for all defense models. This is because ensemble attacks combine the strengths of multiple attacks. It is worth mentioning that the current adversarial robustness leaderboard is created based on the AutoAttack ensemble \citep{croce2020robustbench}. By replacing two of its attacks, our MDE is able to produce even better evaluation with lower robustness in most cases, except a tie on SAT and R-WRN, and 0.05\% worse on UAT. The improvements are more pronounced on the imbalanced gradients models (e.g., the bottom 6 models). This result again verifies the importance of margin exploration in robustness evaluation. 
% Next, we will conduct a more detailed exploration on what defense or attack techniques can cause or circumvent imbalanced gradients by focusing on the imbalanced defense models.

\subsection{Defenses that Cause Imbalanced Gradients}\label{sec:5.2}
% The gradient imbalance ratio of all 12 defense models are illustrated in Appendix \ref{sec:gir}. 
Here, we explore common defense techniques that cause imbalanced gradients by focusing on the 6 defense models that are relatively more susceptible to imbalanced gradients: MMA, Bilateral, Adv-Interp, FeaScatter, Sense, and JARN-AT1.
To avoid other factors introduced by the different attack techniques used by ODI and DLR attacks, here we directly compare the robustness evaluated by the classic PGD attack and that evaluated by our MD attack.
% on the PGD and  robustness of these models and compare it against our MD robustness. 
Note that imbalanced gradients are not the only possible but rather a subtle cause of over-estimated robustness. That is, it is more subtle than obfuscated gradients and needs specific techniques to evade.

\begin{table}[!htb]
% \small
\centering
\caption{Robustness (\%) of WideResNet-34-10 (may be different to those evaluated in Table \ref{tbl:comprehensive}) models trained with/without label smoothing. $z_{max}$, $-z_y$, and $z_{max} - z_y$ denote MD attack with the single loss terms, respectively. The lowest robustness results are \textbf{boldfaced}.
} \label{tbl:ls}
\begin{adjustbox}{width=\linewidth}
\setlength{\tabcolsep}{2mm}{
    \begin{tabular}{lcccccc|ccc}
      \toprule
        \textbf{Defense} & FGSM & CW & PGD & ODI & DLR & MD &
        $z_{max}$ & $-z_y$ & $z_{max} - z_y$ \\
      \midrule 
      Natural & 31.96 & 37.73 & \textbf{0.00} & \textbf{0.00} & \textbf{0.00} & \textbf{0.00} & \textbf{0.00} & 0.02 & \textbf{0.00}\\
        + Label Smoothing & 58.28 & 49.27 & 7.87 & 0.28 & 0.14 & \textbf{0.06} & 0.11 & 0.29 & 0.13 \\ 
          \midrule
      SAT  & 65.92 & 50.56 & 47.25 & 48.86 & 48.87 & \textbf{46.31} & 61.10 & 55.07 & 48.47 \\
      + Label Smoothing & 66.57 & 51.15 & 48.67 & 48.25 & 50.35 & \textbf{46.60} & 52.19 & 48.82 & 48.37 \\
     \bottomrule
    \end{tabular}
}
\end{adjustbox}
\end{table}
      
% \vspace{-2mm}
\begin{figure}[!ht]
\centering
    \includegraphics[width=0.7\linewidth]{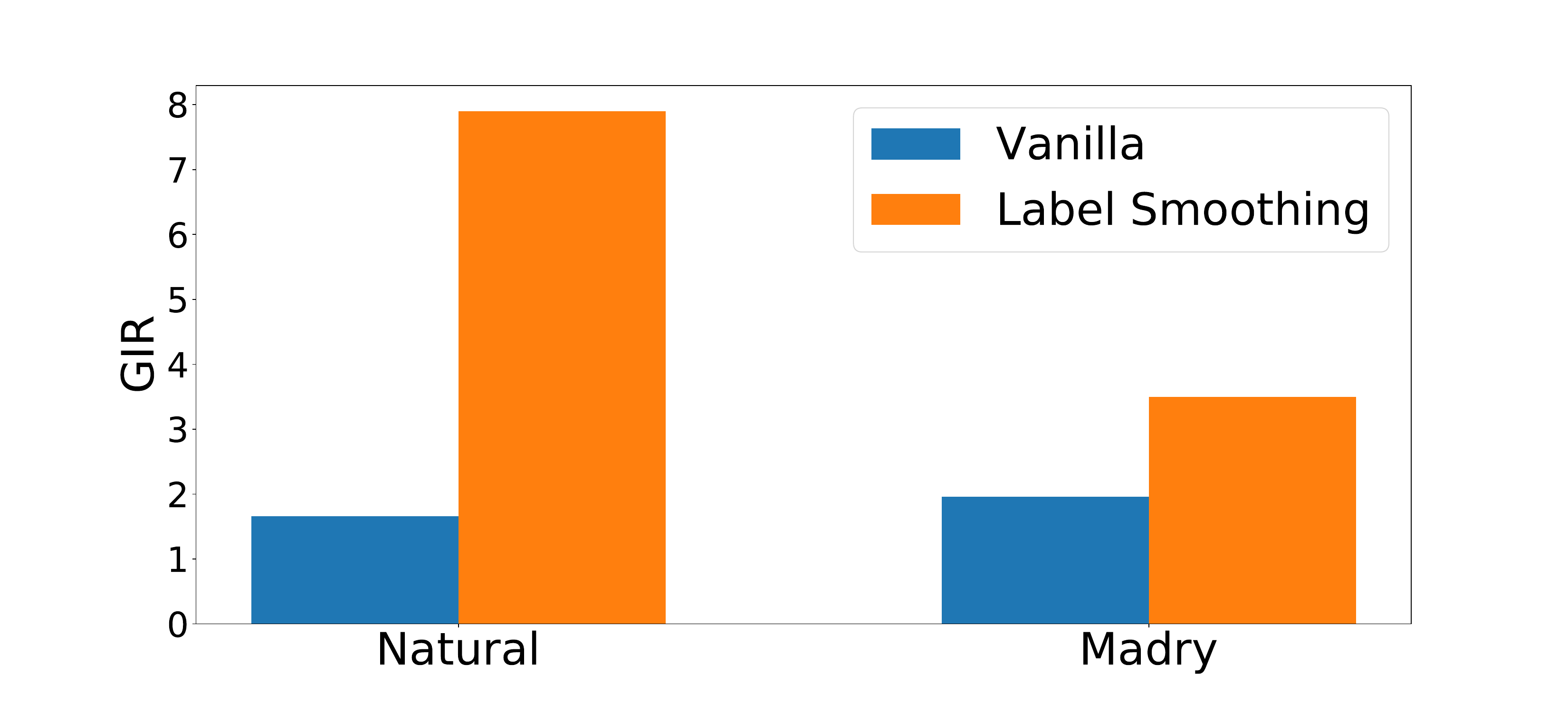}\caption{Gradient Imbalance Ratio (GIR) of models trained with/without label smoothing.}
    \label{fig:ls}
    %   \vspace{-0.2 in}
\end{figure}

\noindent\textbf{Label Smoothing Causes Imbalanced Gradients.}
According to Table \ref{tbl:comprehensive}, the PGD robustness of Adv-Interp, FeaScatter, and Bilateral decreases the most (i.e., $20\% - 27\%$) against our MD attack. This indicates that these defenses have caused the imbalanced gradients problem, as also confirmed by their high GIR values in Fig. \ref{fig:mir}.
All three defenses use label smoothing as part of their training scheme to improve adversarial training, which we suspect is one major cause of imbalanced gradients.
Given a sample $\bm{x}$ with label $y$, label smoothing encourages the model to learn a uniform logit or probability distribution over classes $j \neq y$.
This tends to smooth out the input gradients of $\bm{x}$ with respect to these classes, resulting in smaller gradients.
In order to confirm label smoothing indeed causes imbalanced gradients, we train a WideResNet-34-10 model using natural training (`Natural') and SAT with or without label smoothing (smoothing parameter 0.5). We report their robustness in Table \ref{tbl:ls}, and show their gradient imbalance ratios (GIRs) in Fig. \ref{fig:ls}.
According to the GIRs, adding label smoothing into the training process immediately increases the imbalance ratio, especially in natural training. The PGD robustness of the naturally-trained model also ``increases" to 7.87\%, which is almost zero (0.06\%) according to our MD attack. 
Using smoothed labels in SAT defense also ``increases" PGD robustness by more than 1\%, which in fact is only 0.3\% according to our MD.
Other attacks including CW and DLR are also affected by the label-smoothing effect. This is because CW and DLR are both logit-based attacks, which are more prone to imbalanced gradients. ODI is less sensitive to label smoothing, yet is still not as effective as our MD attack.
These evidences confirm that label smoothing indeed causes imbalanced gradients, leading to overestimated robustness if evaluated by regular attacks like PGD or DLR.
 Fig. \ref{fig:ls} demonstrates that adversarial training can inhibit imbalanced gradients caused by label smoothing to large extent. This is because the adversarial examples used for adversarial training are specifically perturbed to the $j\neq y$ classes, thus helping avoid uniform logits over classes $j \neq y$ to some extent.
 
The last three columns in Table \ref{tbl:ls} show the robustness results when different loss terms are used in MD, in the presence of label smoothing and thus imbalanced gradients. They are consistent with our illustration in Fig. \ref{fig:difference}, i.e., $-z_{y}$ works similarly as $z_{max}--z_{y}$, while $z_{max}$ is the worst. This means that the non-dominant term is more effective when there are imbalanced gradients.

\noindent\textbf{Other Defense Techniques that Cause Imbalanced Gradients.}
The other 3 imbalanced defenses MMA, Sense and JARN-AT1 adopt different defense techniques to ``improve" robustness. MMA is a margin-based defense that maximizes the shortest successful perturbation for each data point. MMA only perturbs correctly-classified examples, and the perturbation stops immediately at misclassification (into a $j \neq y$ class). In other words, MMA focuses on examples that are around the decision boundary (i.e., $\bm{z}_{max} = \bm{z}_y$) between class $y$ and all other classes $j \neq y$. During training, the decision boundary margin is maximized by pulling the boundary away from these examples. This process maximizes the distance to the closest decision boundary (e.g., towards the weakest class) and results in equal distances to all other classes. This tends to generate a uniform prediction over classes $j \neq y$, a similar effect of label smoothing and causes imbalanced gradients.

Similar to MMA, Sense perturbs training examples until a certain loss threshold is satisfied. While in MMA the threshold is misclassification, in Sense, it is the loss value with respect to probability (e.g., $\bm{p}_y=0.7$). This type of training procedures with specific logits or probability distribution regularization has caused the imbalanced gradients problem for both MMA and Sense. Note that, Sense causes more imbalanced gradients than MMA. We conjecture it is because optimizing over a probability threshold is much easier than moving the decision boundary.

JARN-AT1 is also a regularization-based adversarial training method. Different from MMA or Sense, it regularizes the model’s Jacobian (e.g., input gradients) to resemble natural training images. Such explicit input gradient regularizations reduce the input gradients to a much smaller magnitude and only keep the salient part of input gradients. The input gradients associated with other $j \neq y$ classes will be minimized to cause an imbalance to that associated with class $y$.

The above analysis indicates that future defense should avoid using label smoothing, margin smoothing or input gradient regularization techniques, or should be carefully evaluated against our MD attack to check for imbalanced gradients.

\begin{figure}[!htb]
    \centering
    \includegraphics[width=0.7\linewidth]{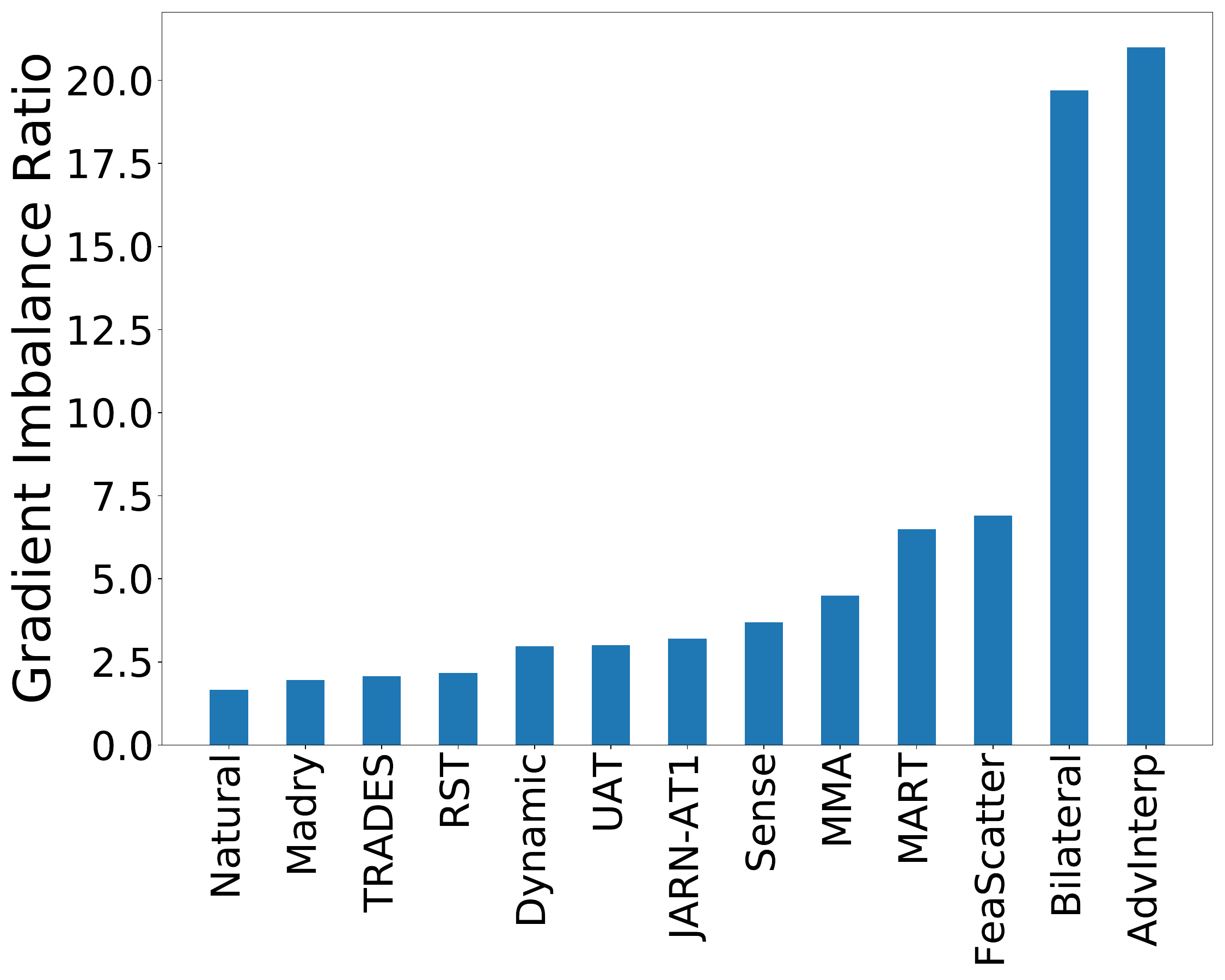}
    \caption{Gradient imbalance ratios (GIRs) of 12 defense models and a naturally trained model (``Natural''). All models are trained on CIFAR-10 dataset. Those defense models that are not susceptible to imbalanced gradients are omitted here.}
    \label{fig:all-mir}
\end{figure}

\noindent\textbf{Correlation between GIR and Robustness.}
According to the GIR scores shown in Fig. \ref{fig:mir} and Fig. \ref{fig:all-mir}, models exhibit high GIR scores (e.g., Adv-Interp, FeaScatter, and Bilateral) are generally more prone to imbalanced gradients and potentially more vulnerable to our MD attack. However, GIR is not a measure of robustness nor should be used as an exact metric to determine whether one defense is more robust than the other. For example, MART demonstrates a higher GIR score than Sense or JARN-AT1, however, according to our MD attack, it is 11\% and 34\% more robust than Sense and JARN-AT1, respectively. 
This is because the GIR score of a model only measures the gradient situation of the model at its current state, which could decrease during the attack as shown in Fig. \ref{fig:rebalance} and \ref{fig:rebalance-attacks}. 
Our MD attack iteratively exploits and circumvents imbalanced gradients during the first attacking stage, thus can produce reliable robustness evaluation at the end.

\subsection{An Attack View of Imbalanced Gradients}\label{sec:5.3}
As shown in Table \ref{tbl:comprehensive}, advanced attacks ODI and DLR are more effective than traditional attacks PGD and CW against imbalanced gradients. Here, we provide more insights into what attack techniques are effective against imbalanced gradients.
Before that, we first show imbalanced gradients are more subtle than obfuscated gradients and cannot be easily circumvented by common techniques like random restarts or momentum.

\subsubsection{Imbalanced Gradients are Different from Obfuscated Gradients}\label{sec:not_obfuscated}
Imbalanced gradients occur when one loss term dominates the attack towards a suboptimal gradient direction, which does not necessarily block gradient descent like obfuscated gradients. 
% Different from obfuscated gradients, imbalanced gradients are still valid gradients, though less effective. 
Therefore, it does not have the characteristics of obfuscated gradients, and cannot be detected by the five checking rules for obfuscated gradients \citep{athalye2018obfuscated}. Here, we test all five rules on the four defense models that exhibit significantly imbalanced gradients: Adv-Interp, FeaScatter, Bilateral, and Sense. Note that all these models were trained and tested on the CIFAR-10 dataset.
% The results are reported in Table \ref{tbl:obfuscated}.

\textbf{One-step attacks outperform iterative attacks}. When gradients are obfuscated, iterative attacks are more likely to get stuck in a local minimum. To test this, we compare the success rate of one-step attack FGSM and iterative attack PGD in Table \ref{tbl:obfuscated}. We see that PGD outperforms FGSM consistently on all four defense models with no obvious sign of obfuscated gradients.

\textbf{Unbounded attacks do not reach 100\% success. Increasing the distortion bound does not increase success.} Larger perturbation bound gives the attacker more ability to attack. So, if gradients are not obfuscated, an unbounded attack should reach a 100\% success rate. To test this, we run an ``unbounded" PGD attack with $\epsilon=255/255$. As shown in Table \ref{tbl:obfuscated}, all models are completely broken by this unbounded attack, i.e., the over-estimated robustness is caused by a more subtle effect than obfuscated gradients.

\textbf{Black-box attacks are better than white-box attacks.} If a model is obfuscating gradients, it should fail to provide useful gradients in a small neighborhood. Therefore, using a substitute model should be able to evade the defense, as the substitute model was not trained to be robust to small perturbations. To test this, we run a black-box transfer PGD attack on naturally trained substitute models.
We find that all four defenses are robust to transferred attacks (``Transfer" in Table \ref{tbl:obfuscated}).
We also attack the four defense models using gradient-free attack SPSA \citep{Uesato2018spsa}. For SPSA, we use a batch size of 8192 with 100 iterations and run on 1000 randomly selected CIFAR-10 test images.
We confirm that SPSA cannot degrade its performance. None of these results indicate obfuscated gradients.

\textbf{Random sampling finds adversarial examples.} Brute force random search within some $\epsilon$-neighbourhood should not find adversarial examples when gradient-based attacks do not. 
Following \citep{athalye2018obfuscated}, we choose 1000 test images on which PGD fails.
We then randomly sample $10^5$ points for each image from its $\epsilon=8/255$-ball region and check if any of them are adversarial.
The results (i.e., ``Random" in Table \ref{tbl:obfuscated}) show that random sampling cannot find adversarial examples when PGD does not.

All the above test results lead to one conclusion that the robustness of the four defenses is not a result of obfuscated gradients. This indicates that imbalanced gradients do not share the characteristics of obfuscated gradients, and thus cannot be detected following the five test principles for obfuscated gradients. 
Therefore, imbalanced gradients should be addressed independently for more reliable robustness evaluation.

\begin{table*}[!htb]
\small
\centering
  \caption{Test of obfuscated gradients for four defense models that have significant imbalanced gradients following \citep{athalye2018obfuscated}: attack success rate (\%) of different attacks. The results indicate no sign of obfuscated gradients.}
\begin{adjustbox}{width=\linewidth}
\setlength{\tabcolsep}{2mm}{
\begin{tabular}{l|ccccccc}
  \toprule
    \textbf{Defense} & FGSM & PGD & Unbounded & Transfer & SPSA & Random  \\
  \midrule
    Adv-Interp \citep{zhang2020adversarial} & 23.06 & 27.52 & 100.00 & 10.89 & 24.80 & 0.00\\
    FeaScatter \citep{zhang2019featurescatter}  & 22.60  & 31.36 & 100.00 & 11.11 & 28.20 & 0.00\\
    Bilateral \citep{wang2019bilateral} & 28.90 & 39.05 & 100.00 &  9.23 & 36.00 & 0.00\\
    % CLP \citep{wang2019bilateral} & 21.82 & 51.74 & 100.00 &  20.94 & 43.50 & 0.00\\
    Sense \citep{kim2020sensible} & 27.29 & 40.14 & 100.00 & 9.90 & 37.90 & 0.00 \\
 \bottomrule
\end{tabular}
}
\end{adjustbox}
  \label{tbl:obfuscated}
\end{table*}

\subsubsection{Momentum, Random Restart Cannot Circumvent Imbalanced Gradients} \label{sec:restart}
As we discussed in Section \ref{sec:toy}, random restarts can potentially increase the probability of finding an adversarial example. Momentum method is another way to help escape overfitting to local gradients \citep{sutskever2013importance}.
Here, we test their effectiveness in circumventing imbalanced gradients.
For the random restart, we run a 400-step PGD attack with 100 restarts ($\text{PGD}^{100\times400}$). For momentum, we use momentum iterative FGSM (MI-FGSM) \citep{dong2018boosting} with 40 steps, 2 restarts, and momentum 1.0. For both attacks, we set $\epsilon=8/255$ and step size $\alpha=2/255$.
Note that, we removed samples that were successfully perturbed from the batch, thus only restarting samples that are not successful. This slightly improves the performance and shows it is stronger than PGD without restarts.
We apply the two attacks on 1000 randomly chosen CIFAR-10 test images and report the robustness in Table \ref{tbl:ifgsm} for the four defense models checked in Section \ref{sec:not_obfuscated}.
Compared to traditional PGD with 40 steps, the robustness can indeed be decreased by $\text{PGD}^{100\times400}$ except on Bilateral, a consistent observation with our analysis in Section \ref{sec:toy} that more restarts can lower model accuracy. However, the robustness is still highly overestimated compared to that of our MD attack. This indicates that imbalanced gradients can exist in wide-spanned input regions, resulting in a low probability of random restart to find successful attacks. To our surprise, MI-FGSM performs even worse than traditional PGD. On three defense models (
i.e., Adv-Interp, FeaScatter, and Sense), it produces even higher robustness than PGD. 
This implies that accumulating velocity in the gradient direction can make the overfitting even worse when there are imbalanced gradients. This again confirms that the imbalanced gradients problem should be explicitly addressed.

\begin{table}[!htb]
\small
\centering
  \caption{Robustness (\%) of four defense models that have significant imbalance gradients against PGD, $\text{PGD}^{100\times400}$, MI-FGSM, and our MD attacks. The best results are \textbf{boldfaced}.}
  \label{tbl:ifgsm}
\setlength{\tabcolsep}{2mm}{
\begin{tabular}{l|cc|cc}
  \toprule
    \textbf{Defense} & PGD & MD (ours) & $\text{PGD}^{100\times400}$ & MI-FGSM   \\
  \midrule
    Adv-Interp & 72.48 & \textbf{45.07} & 71.64 & 73.25 \\
    FeaScatter  & 68.64 & \textbf{42.16} & 67.00 & 70.79 \\
    % CLP & 48.26 & \textbf{0.74} & 20.60 & 55.53 \\
    Sense & 59.86 & \textbf{39.91} & 58.51 & 62.41 \\
    Bilateral & 60.95 & \textbf{39.39} & 59.61 & 51.52 \\
    
 \bottomrule
\end{tabular}
}
\end{table}

\subsubsection{What Can Help Circumvent Imbalanced Gradients?}
\begin{figure}%[!ht]
\vskip -0.2in
	\centering
	\begin{subfigure}{0.3\linewidth}
	    \centering
		\includegraphics[width=\textwidth]{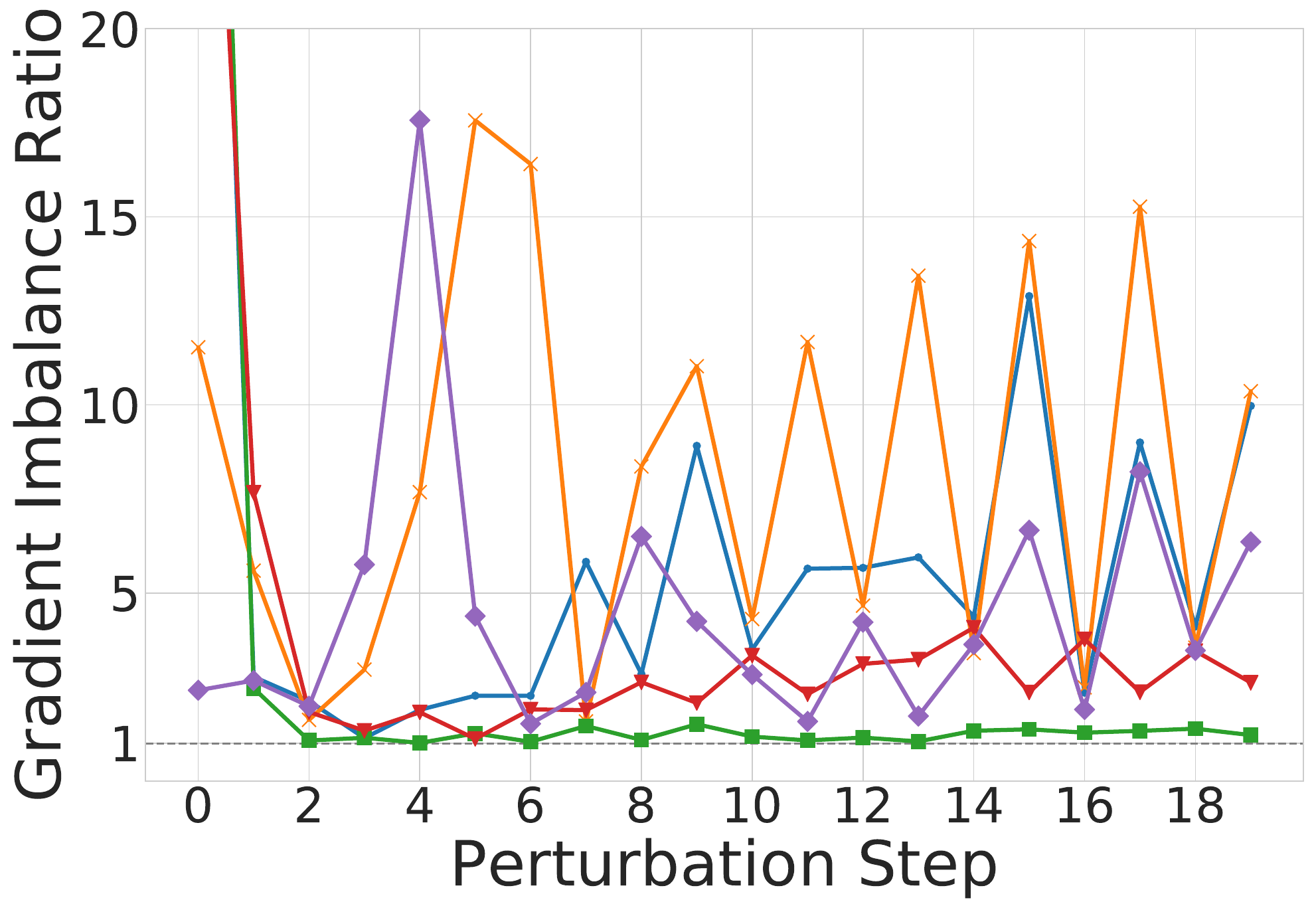}
		\caption{ODI}
		\label{fig:odi}
	\end{subfigure}
    \hspace{0.2cm}
	\begin{subfigure}{0.3\linewidth} 
	    \centering
		\includegraphics[width=\textwidth]{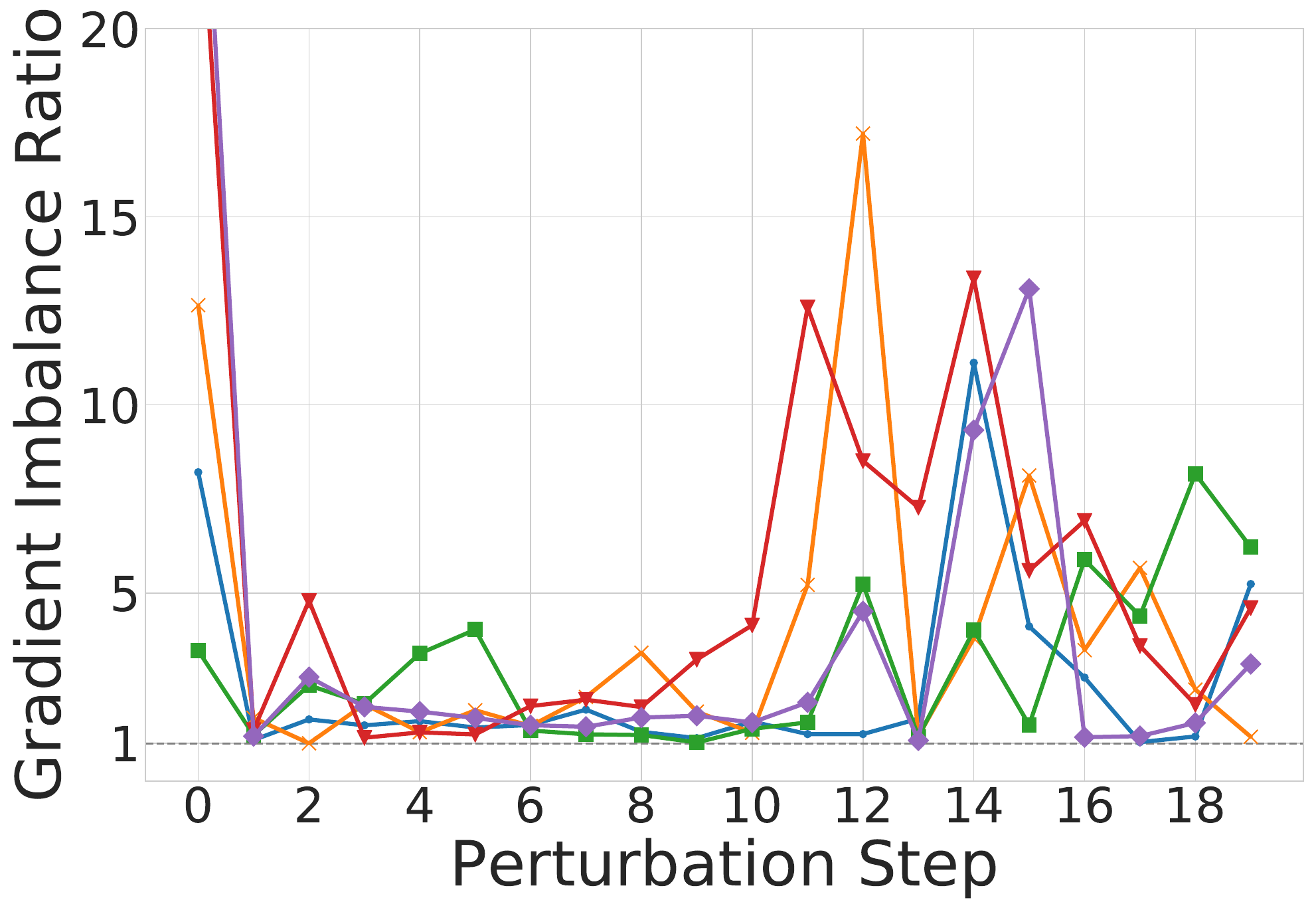}
		\caption{FAB}
		\label{fig:fab}
	\end{subfigure}
    \hspace{0.2cm}
	\begin{subfigure}{0.3\linewidth}
	    \centering
		\includegraphics[width=\textwidth]{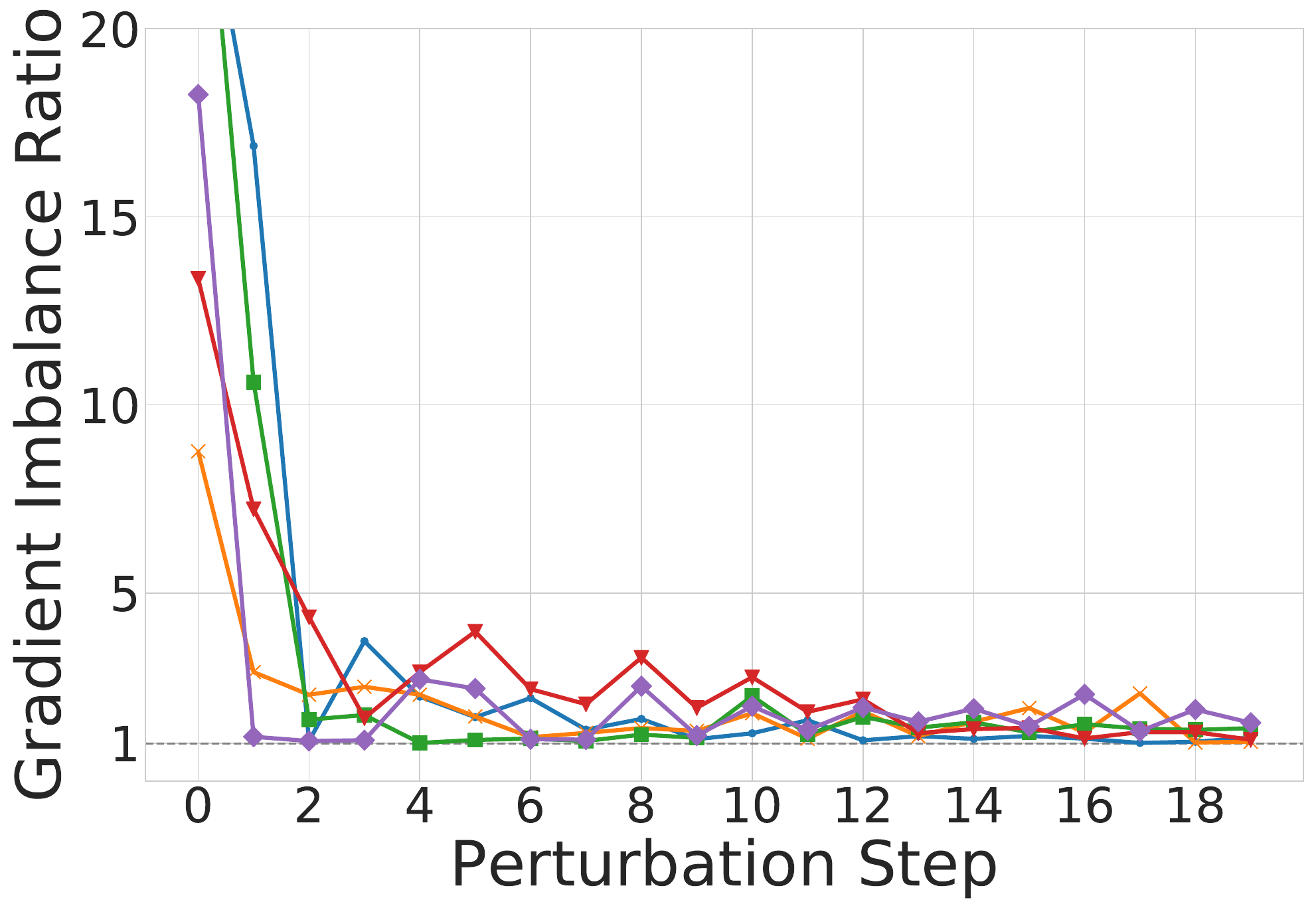}
		\caption{MD (Ours)}
		\label{fig:md}
	\end{subfigure}
  \caption{Gradient imbalance ratio at the first 20 steps of ODI (a), FAB (b) and our MD (c) attacks on the AdvInterp model for 5 randomly selected CIFAR-10 test images.}
  \vskip -0.2in
\label{fig:rebalance-attacks}
\end{figure}

\noindent\textbf{Logits Diversified Initialization.} ODI randomly initializes the perturbation by adding random weights to logits at its first 2 steps.
The random weights change the gradient magnitude, and thus can also mitigate imbalanced gradients, as shown in Fig. \ref{fig:odi}.
However, the initialization can only help with the first two attack steps, and the imbalance ratio fluctuates drastically in the following steps.
Our MD attack provides a more direct and efficient exploration of imbalance gradients, thus can maintain a low imbalance ratio even after the first few steps (see Fig. \ref{fig:md}). In Table \ref{tbl:comprehensive}, our MD attack was also found to be more effective than ODI.

\noindent\textbf{Exploration Beyond the $\epsilon$-ball.}
% AutoAttack is an ensemble of four attacks: two Auto-PGD attacks and two existing attacks FAB and Square. 
By inspecting the individual attacks in AutoAttack, we found that the exploration technique used by the FAB attack is also effective against imbalanced gradients to some extent.
FAB first finds a successful attack using unbounded perturbation size, then minimizes the perturbation to be within the $\epsilon$-ball. As shown in Fig. \ref{fig:fab}, the first few steps of exploration outside the $\epsilon$-ball can effectively avoid imbalanced gradients. This is also why our MD attacks use a large step size in the first stage.
However, the imbalance ratio tends to increase when FAB attempts to minimize the perturbation (steps 10 - 16). We believe FAB can be further improved following a similar strategy to our margin decomposition.
% We also find that optimization techniques like momentum that can effectively avoid local minima are not effective against imbalanced gradients, as shown in Appendix \ref{sec:restart}.

\noindent\textbf{Circumventing Imbalanced Gradients Improves Black-box Attacks.} Here we show gradient estimation based black-box attacks can also benefit from our MD method when there are imbalanced gradients. We take SPSA as an example and use the two-stage losses of our MD attack for SPSA. This version of SPSA is denoted as SPSA+MD. 
For both SPSA and SPSA+MD, we use the same batch size of 8192 with 100 iterations. Since black-box attack is quite time-consuming, we only run on 1000 randomly selected CIFAR-10 test images.
% Note that we use the same number of perturbation steps and step size for both SPSA and SPSA+MD. 
The attack success rates on Adv-Interp, FeaScatter, and Sense models are reported in Table \ref{tab:black-box}.
Compared to SPSA, SPSA+MD can lower the robustness by at least 10.9\%. This indicates that imbalanced gradients also have a negative impact on back-box attacks, and our method can be easily applied to produce more query-efficient and successful black-box attacks.
% For gradient estimation based black-box attacks like SPSA, using our MD attacks can also help the attack to escape imbalanced gradients, leading to more queries-efficient and successful attacks. 

\begin{table}[!htb]
\small
\centering
% \vspace{-3mm}
  \caption{Attack success rate (ASR, \%) of the SPSA attack with or without our MD losses on three defense models. `$\uparrow$' marks the ASR boost by MD. The best results are \textbf{boldfaced}.}\label{tab:black-box}
\begin{tabular}{l|ccc}
  \toprule
    \textbf{Attack} & Adv-Interp & FeaScatter&Sense\\
  \midrule
    SPSA & 24.80 & 28.20&37.90 \\
    SPSA+MD &  \textbf{40.30} ($\uparrow$15.5) & \textbf{45.60} ($\uparrow$17.4) & \textbf{48.80} ($\uparrow$10.9) \\
 \bottomrule
\end{tabular}
% \vspace{-3mm}
\end{table}

\subsection{Ablation and Parameter Analysis of MD Attack}\label{sec:ablation}
In this section, we provide a more detailed analysis of our proposed MD attack via an ablation study and a parameter analysis. The ablation study focuses on the two attacking stages of MD, while the parameter analysis focuses on the perturbation-related parameters including the number of steps and initial step size.

\subsubsection{Ablation Study}
Here, we investigate two influential factors to our MD attack: 1) the second attacking stage, and 3) the stage order. We use AdvInterp as our target model and conduct the following attack experiments on CIFAR-10 test data.

% \noindent\textbf{Initialization Method.} We compare the success rates of our MD attacks using random initialization versus the opposite direction initialization (see Algorithm \ref{alg:ld-pgd} and \ref{alg:MD-MT}). 
% The results are reported in Table \ref{tab:ablation}.
% As can be observed, the opposite direction initialization demonstrates a clear advantage over random initialization. 
% Particularly, for MD attack, using opposite direction initialization can improve the attack success rate by 8\%, while for MD-MT attack, the success rate can also be improved. 
% a improvement of 1.34\% can be achieved by opposite direction initialization. This indicates that 
% Particularly, for PGD, MT attacks, using opposite direction initialization can improve the success rate by 16\% and 13\% respectively. 

\noindent\textbf{The Second Attacking Stage.}
We further investigate the importance of the second stage of attacking with the full margin loss in our MD and MD-MT attacks.
The attack success rates with or without the second stage are also reported in Table \ref{tab:ablation}.
It shows that attacking the full margin loss via the second stage can increase the success rate, for both MD and MD-MT. This verifies the importance of the second stage for generating the strongest attacks.

\noindent\textbf{Ordering of the Two Stages.}
To verify that the ordering of the two stages is suitable for MD attacks, we evaluate a new version of our MD attacks with the two stages switched: the first stage optimizes the full margin loss and the second stage explores the individual loss terms. The results are also reported in Table \ref{tab:ablation} (the last two columns). As can be observed, MD attacks become less effective when the two stages are switched, even compared to that without the second stage. 
This indicates that the imbalanced gradients should be addressed first before producing a reliable robustness evaluation. 

\begin{table*}[!htb]
\renewcommand{\arraystretch}{1.1}
\centering
\small
  \caption{Attack success rates (\%) of our MD and MD-MT attacks 1) with or without the second stage, and 2) with or without the two stages being switched.  Experiments are conducted on the defense model AdvInterp and CIFAR-10 dataset. The best results are \textbf{boldfaced}.}
  \label{tab:ablation}
  \begin{tabular}{c|cc|cc}
    \toprule
    \multirow{2}{*}{Attacks} & \multicolumn{2}{c}{Second Stage} &  \multicolumn{2}{c}{Switching Stage}\\  & Without & With & Yes & No \\
    \hline
    MD & 44.53 & \textbf{45.18} & 43.17 & \textbf{45.18} \\
    MD-MT & 52.56 & \textbf{52.71} & 51.47
 & \textbf{52.71} \\
    \bottomrule
  \end{tabular}
\end{table*}

\subsubsection{Parameter Analysis}\label{sec:parameter_analysis}
We further investigate the sensitivity of our MD attack to two parameters: 1) the number of perturbation steps, and 2) the initial step size. Here, we focus on the first attacking stage as the second stage is similar to the PGD attack, which has been investigated in \citep{wang2019convergence}.

\begin{figure}[h]
    \centering
    \subcaptionbox{\label{fig:numsteps}}{\includegraphics[width=0.49\textwidth]{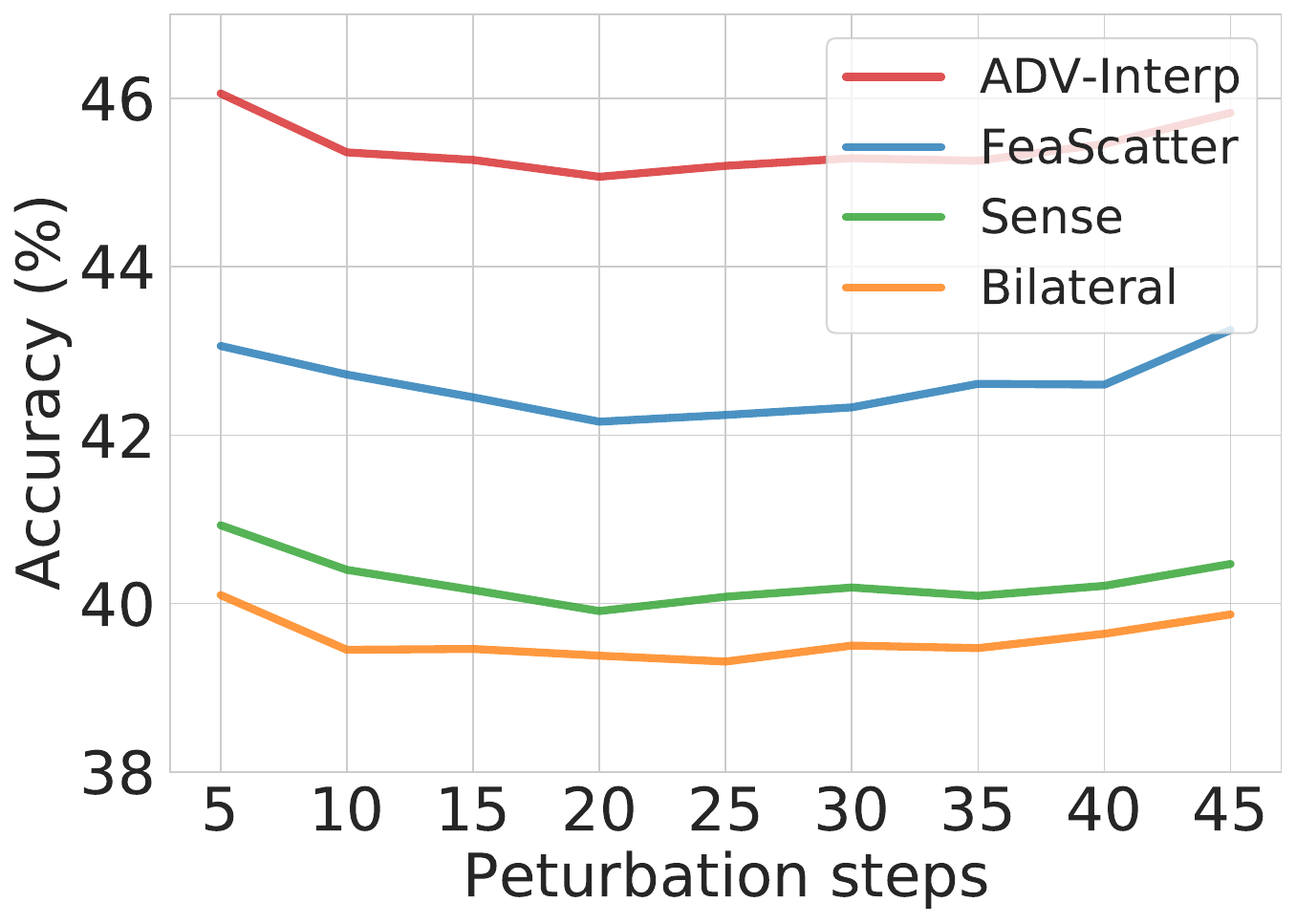}}%
    \hfill
    \subcaptionbox{\label{fig:stepsize}} {\includegraphics[width=0.49\textwidth]{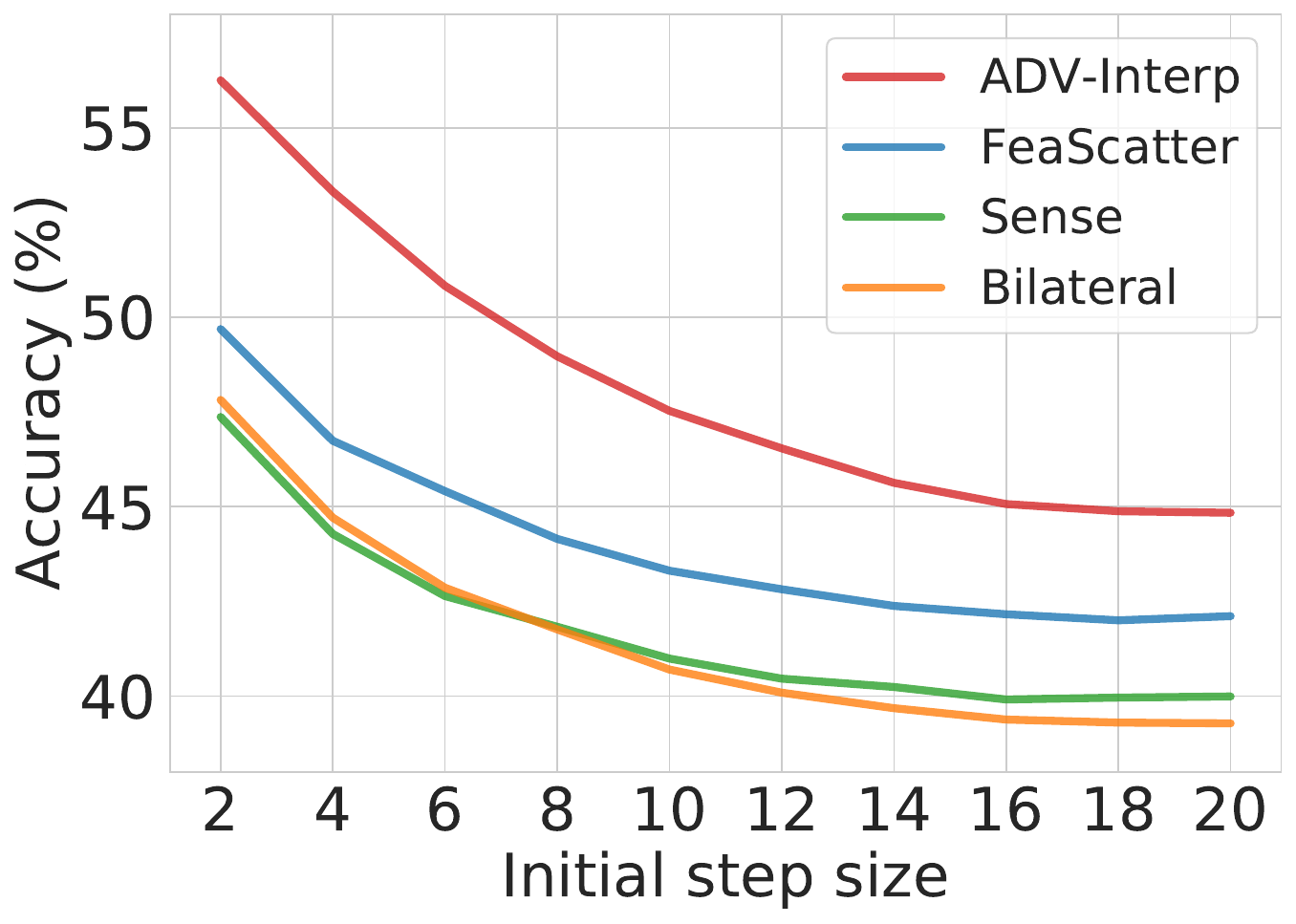}}
    \caption{Parameter analysis: the accuracies (robustness) of four defense models under our MD attack with a different (a) number of steps, or (b) initial step sizes for the first stage.}
\end{figure}

\textbf{Number of Steps for the First Stage.}
Here, we fix the total number of steps for the two stages to $K=100$ and vary the steps allocated for the first stage.
Note that MD attack will reduce to the regular PGD attack if the step of its first stage is set to 0.
Here, we vary the steps for the first stage from 5 to 50 in granularity of 5. The initial step size is fixed to $\alpha=2\epsilon$ and gradually decreased to 0 via cosine annealing.
The robustness of four defense models including Bilateral, Sense, Adv-Interp, and FeaScatter are illustrated in Fig. \ref{fig:numsteps}.
As can be observed, the effectiveness of our MD attack tends to drop at both ends, and the best performance (lowest evaluated robustness) is achieved at 20, except for Bilateral (which is 25). Therefore, we suggest using 20 steps as the optimal choice for the first stage. 

\textbf{Initial step Size for the First Stage.} We vary the initial step size used in the first stage from 2/255 to 20/255 in a granularity of 2/255. The initial step size will decrease to 0 with the cosine annealing. Following the above experiment, here we fix the number of steps for the first stage to 20. The evaluated robustness (or model accuracy on the generated attacks) of the four defense models are illustrated in Fig. \ref{fig:stepsize}.
A clear improvement of using a large initial step size can be observed. Based on the trend, we suggest using a relatively large ($\alpha\in[2\epsilon,3\epsilon)$) initial step size to help circumvent imbalanced gradients during the first stage.

\section{Conclusion}
In this paper, we identified a subtle situation called \emph{Imbalanced Gradients}, where existing attacks may fail to produce the most accurate adversarial robustness evaluation. We proposed a new metric to investigate the imbalanced gradients problem in current defense models.
We also proposed a new attack called Margin Decomposition (MD) attack to leverage imbalanced gradients via a two-stage attacking process. The multi-targeted and ensemble version of MD attacks were also introduced to generate the strongest attacks.
By re-evaluating 24 defense models proposed since 2018, we found that 11 of them are susceptible to imbalanced gradients to some extent and their robustness evaluated by the best standalone attack can be further reduced for more than 1\% by our MD attack. We identified a set of possible causes of imbalanced gradients, and effective countermeasures.
Our results indicate that future defenses should avoid causing imbalanced gradients to achieve more reliable adversarial robustness.

\section*{Acknowledgments}
This work was supported in part by the National Key R\&D Program of China (Grant No. 2021ZD0112804), National Natural Science Foundation of China (Grant No. 62276067 and 62032006), and Shanghai Science and Technology Committee (Grant No. 20511101000). This research was undertaken using the LIEF HPC-GPGPU Facility hosted at the University of Melbourne. This Facility was established with the assistance of LIEF Grant LE170100200.

\section*{Broader Impact}
Our work highlights a subtle pitfall in adversarial robustness evaluation and provides a more accurate evaluation method. It can benefit different application domains to examine the robustness of deep learning models against adversarial examples. The failure of our method may lead to less accurate evaluations, which should be carefully examined with other methods. 

\bibliography{ref}
\bibliographystyle{plainnat}

\end{document}